\documentclass[twocolumn,final]{elsarticle}

\usepackage{ycviu}
\usepackage{framed,multirow}

\usepackage{amssymb}
\usepackage{latexsym}

\usepackage{url}
\usepackage{xcolor}
\definecolor{newcolor}{rgb}{.8,.349,.1}

\usepackage{multirow}
\usepackage{mathptmx}
\usepackage{mathtools}		
\usepackage[normalem]{ulem}	
\usepackage{comment}
\usepackage{color}
\usepackage{subfigure}
\usepackage[autostyle]{csquotes}

\definecolor{R}{rgb}{0.9,0.3,0.3} 
\definecolor{G}{rgb}{0.3,0.9,0.3} %
\definecolor{B}{rgb}{0.3,0.3,0.9} 

\journal{Computer Vision and Image Understanding}

\begin{document}

\ifpreprint
  \setcounter{page}{1}
\else
  \setcounter{page}{1}
\fi

\begin{frontmatter}

\title{Registration-Free Face-SSD: Single Shot Analysis of Smiles, Facial Attributes, and Affect in The Wild}

\author[1]{Youngkyoon \snm{Jang}\corref{cor1}} \cortext[cor1]{Corresponding author: 
  Tel.: +44-(0)752-214-2643;}
\ead{youngkyoon.jang@bristol.ac.uk}
\author[2]{Hatice \snm{Gunes}}
\author[3]{Ioannis \snm{Patras}}

\address[1]{University of Bristol, 1 Cathedral Square, Trinity Street, Bristol BS1 5DD, UK}
\address[2]{University of Cambridge, William Gates Building, 15 JJ Thomson Avenue, Cambridge CB3 0FD, UK}
\address[3]{Queen Mary University of London, Mile End Road, London E1 4NS, UK}

\received{1 May 2013}
\finalform{10 May 2013}
\accepted{13 May 2013}
\availableonline{15 May 2013}
\communicated{S. Sarkar}

\begin{abstract}
In this paper, we present a novel single shot face-related task analysis method, called Face-SSD, for detecting faces and for performing various face-related (classification / regression) tasks including smile recognition, face attribute prediction and valence-arousal estimation in the wild. Face-SSD uses a Fully Convolutional Neural Network (FCNN) to detect multiple faces of different sizes and recognise / regress one or more face-related classes. Face-SSD has two parallel branches that share the same low-level filters, one branch dealing with face detection and the other one with face analysis tasks. The outputs of both branches are spatially aligned heatmaps that are produced in parallel -- therefore Face-SSD does not require that face detection, facial region extraction, size normalisation, and facial region processing are performed in subsequent steps. Our contributions are threefold: 1) Face-SSD is the first network to perform face analysis without relying on pre-processing such as face detection and registration in advance -- Face-SSD is a simple and a single FCNN architecture simultaneously performing face detection and face-related task analysis -- those are conventionally treated as separate consecutive tasks; 2) Face-SSD is a generalised architecture that is applicable for various face analysis tasks without modifying the network structure -- this is in contrast to designing task-specific architectures; and 3) Face-SSD achieves real-time performance ($21$ FPS) even when detecting multiple faces and recognising multiple classes in a given image ($300\times300$). Experimental results show that Face-SSD achieves state-of-the-art performance in various face analysis tasks by reaching a recognition accuracy of $95.76\%$ for smile detection, $90.29\%$ for attribute prediction, and Root Mean Square (RMS) error of $0.44$ and $0.39$ for valence and arousal estimation. 
\end{abstract}


\end{frontmatter}


\section{Introduction}
\label{sec: introduction}
\begin{figure*} [t!]
\centering
    \subfigure[Smile Recognition]
    {
    	\label{subfig:smile_ex}
      	\includegraphics[width=0.32\linewidth]{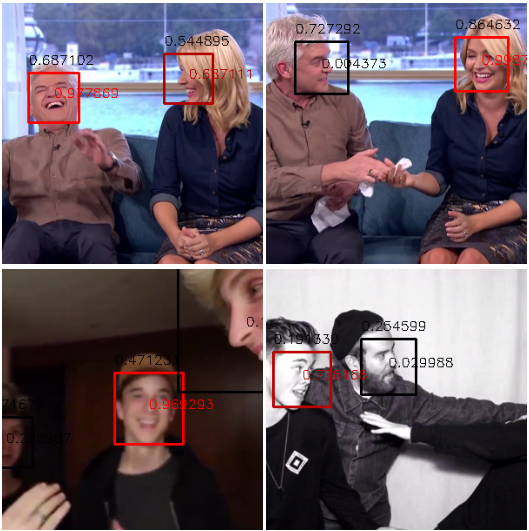}
	}
    \subfigure[Facial Attribute Prediction]
    {
      \label{subfig:attr_ex}
      \includegraphics[width=0.32\linewidth]{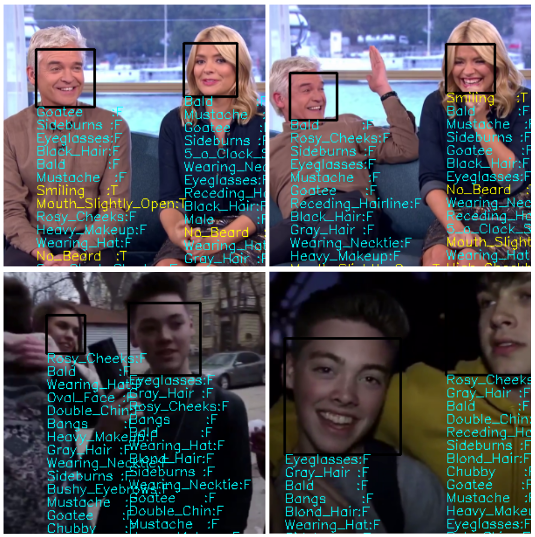}
    }
    \subfigure[Valence-Arousal Estimation]
    {
      \label{subfig:v_a_ex}
      \includegraphics[width=0.32\linewidth]{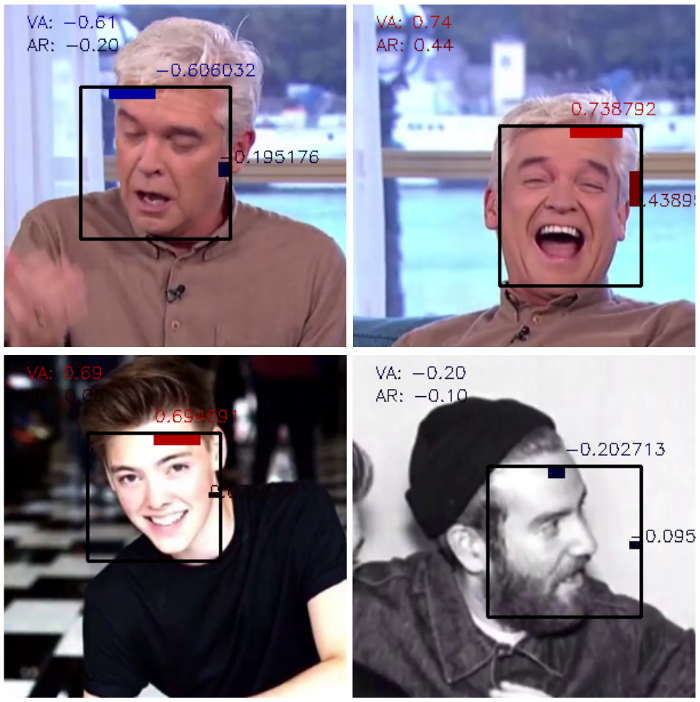}
    }
\caption{{Our system, which we refer to here as Face-SSD, detects faces and smiles, recognises facial attributes, and predicts affect along the valence and arousal dimensions, in the wild. {\subref{subfig:smile_ex}} When detected faces are determined as smiling faces, the colour of the black bounding box changes to red. The probability that appears at the top of the box indicates the face confidence score, and the one appearing in the middle of the box is the smile confidence score. The intensity of red corresponds to the level of confidence. {\subref{subfig:attr_ex}} $40$ attribute labels that are sorted in descending order (using prediction score) are displayed below the corresponding face bounding box. Attributes classified as ``True'' are displayed in yellow text, or displayed in another colour otherwise. For sorting, we use a modified value ($1.0$ $-$ predicted score) for the attributes that are classified as ``False''. {\subref{subfig:v_a_ex}} Horizontal and vertical bars indicate the degree of valence (VA) and arousal (AR), respectively. A bar starts in the middle (representing a value of zero) and ends at the corner of the bounding box along both positive (right / top) and negative (left / bottom) directions. The intensity of the colour (red for positive, blue for negative) corresponds to the level of the predicted score between $-1$ and $1$. According to the content of the dataset, Face-SSD has been trained for a limited range of face sizes (see Sec. \ref{subsec: V-A estimation performance} for details). (Best viewed in colour)}}
\label{fig:teaser}
\end{figure*}

Face analysis is one of the most studied areas in various research communities including Computer Vision (CV) and Affective Computing (AC). Cutting edge results are constantly obtained for various face-related analysis and recognition tasks including face detection \citep{conf/ICCV/KZhang17, conf/ICCV/SZhang17, jour/neurocomp/SZhang18}, face recognition \citep{conf/ICCV/Wu17}, expression recognition \citep{conf/CVPR/Li17}, valence-arousal estimation \citep{jour/IVC/Kossaifi17}, action unit detection \citep{jour/TIP/Zeng16, conf/CVPR/Li17}, face attribute recognition \citep{conf/ICCV/Liu15, conf/AAAI/Hand17}, age estimation \citep{conf/CVPR/Chen17, conf/CVPRW/Hsu17, conf/FG/Agustsson17}, landmark detection \citep{conf/CVPR/Lv17, conf/ICCVW/Shen17} and face alignment \citep{jour/IJCV/Jourabloo17}. However, in order to get the best performance, recent studies design specific architectures for each individual face analysis task. Although some works propose unified frameworks for handling multiple face-related tasks {\citep{conf/CVPR/Wu17, conf/CVPRW/Chang17, conf/FG/Ranjan16}}, several open issues remain yet to be explored:

\begin{itemize}
\item {\textbf{Unconstrained conditions:}} Most of the existing approaches require a detected and normalised face input.
\item {\textbf{Scalability:}} Most methods design separate networks for different tasks. However, networks that are specifically designed to maximise the performance for certain tasks cannot be easily adapted to do other types of face analysis tasks.
\item {\textbf{Real-time performance:}} Existing methods do not achieve real-time performance because they require time-consuming preprocessing steps such as face detection and registration before performing face analysis.
\end{itemize}

\begin{figure*} [t!]
\centering
      \includegraphics[width=0.95\linewidth]{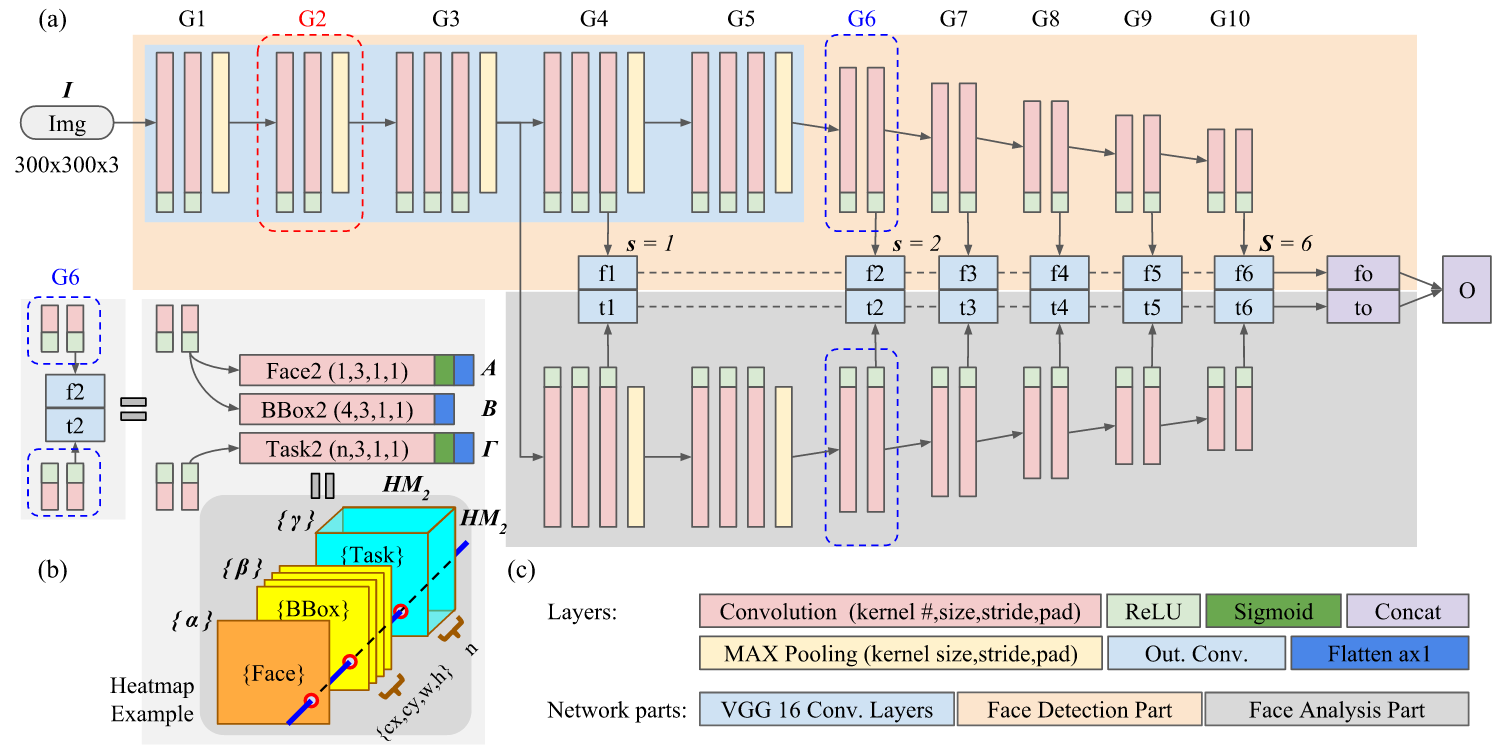}\\
      \includegraphics[width=0.95\linewidth]{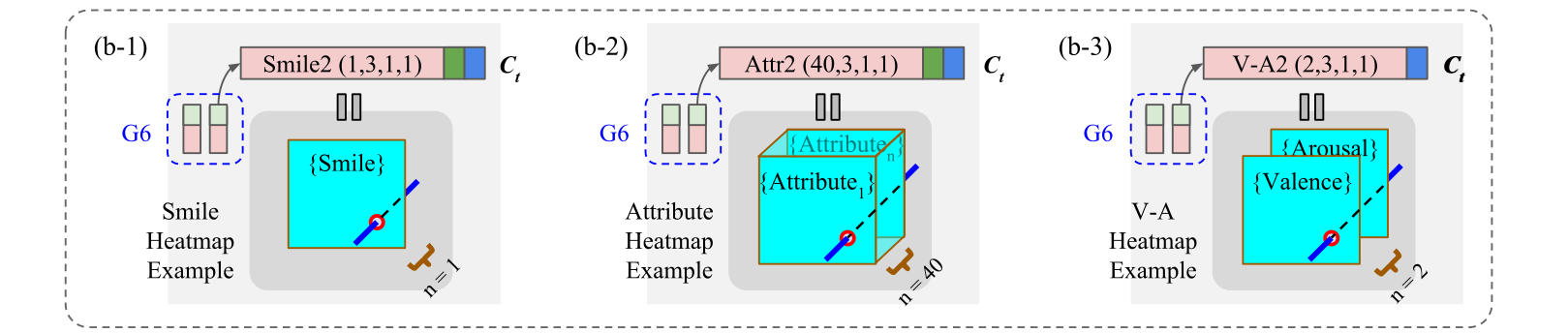}
\caption{{The architecture of Face-SSD. (a) The entire architecture of Face-SSD consisting solely of convolution and pool layers. (b) Example of the concatenated output convolution layer for the second scale ($s=2$) that produces a heatmap volume. (b-1) through (b-3) show how to modify Face-SSD for binary classification (smile recognition), multi-class recognition (facial attribute prediction), and multi-task regression (valence-arousal estimation). (c) Legend for layers and parts of Face-SSD.}}
\label{fig:basic_structure}
\end{figure*}

In order to address the above mentioned challenges, we propose Face-SSD, a network that performs simultaneously face detection and one or more face analysis tasks (see Fig. {\ref{fig:teaser}}) in a single architecture. Face-SSD aims to not only detect faces in a given colour image (upper part in Fig. {\ref{fig:basic_structure}} (a)), but also to perform several other face analysis tasks (lower part in Fig. {\ref{fig:basic_structure}} (a)) associated with the detected faces. Similar to the SSD used for object detection {\citep{conf/ECCV/Liu16}}, the proposed Face-SSD uses a pre-trained VGG16 network {\citep{conf/ICLR/Simonyan115}} to extract low level features as shown in Fig. {\ref{fig:basic_structure}} (a) [G1:G5]. Then, multi-scaled convolution layers are added after the convolutional layers of the VGG16 to perform both classification (face classification and face analysis task) and regression (bounding box localisation) tasks (see Fig. {\ref{fig:basic_structure}} (a) [G6:G10]). To the best of our knowledge, Face-SSD is the first single face network that can handle several face analysis tasks without a pre-normalisation step.

The proposed architecture is trained and evaluated using well-known benchmark datasets for face detection (AFLW {\citep{conf/IWBFIAT/Koestinger11}}), smile recognition (GENKI-4K {\citep{GENKI_DB}}, CelebA {\citep{conf/ICCV/Liu15}}), facial attribute prediction (CelebA {\citep{conf/ICCV/Liu15}}) and valence-arousal estimation (AffectNet {\citep{jour/TAC/Mollahosseini17}}). As discussed in Sec. \ref{subsec:SmileNet_Insignts}, we first obtain a set of matched default boxes as proposed by Liu et al. {\citep{conf/ECCV/Liu16}}. Then, we train the Face-SSD by optimising multiple losses (associated with face classification, bounding box regression, and a face-related task such as smile recognition, multiple facial attributes prediction or valence-arousal estimation). We adopt data augmentation and Hard Negative Mining (HNM) strategies, and achieve state-of-the-art or very competitive performance in various face analysis tasks without modifying the structure and while maintaining real-time performance.

The main contributions of our work are three-fold:

\begin{enumerate}
\item {\textbf{Unconstrained processing:}} Face-SSD does not rely on a pre-normalisation step, it requires neither face detection nor registration in advance. Most of the existing approaches to face analysis require a cropped or normalised face in advance.
\item {\textbf{Universal architecture:}} Face-SSD can be applied to most face analysis tasks with a simple modification (the number of final prediction channels), and achieve state-of-the-art or very competitive results. Most of the existing approaches use separate networks.
\item {\textbf{Real-time processing:}} Face-SSD can be trivially extended to perform several face analysis tasks at negligible additional processing time.
\end{enumerate}

The remainder of the paper is organised as follows: In Section \ref{sec: related work}, related work in face analysis that rely on registration, require task-specific model design, or handle multiple tasks is reviewed. In Section \ref{sec: method}, we present the proposed Face-SSD framework and explain how to apply Face-SSD to several face applications. 
Experimental results of the applications using the proposed Face-SSD are provided in Section \ref{sec: experiments}. Finally, conclusions are drawn and discussed in Section \ref{sec: conclusion}.


\section{Related Work}
\label{sec: related work}

{\textbf{General pipeline for facial analysis.}} Sariyanidi et al. {\citep{jour/TPAMI/Sariyanidi15}} discusses the state-of-the-art methods for face registration, representation, dimensionality reduction and recognition, which are the common components of a generic pipeline for performing automatic facial affect analysis. Depending on the target application, the generic pipeline might have to be changed to some degree. Nonetheless, the first two steps of face localisation and 2D / 3D registration steps have been necessary for most of the face analysis tasks such as smile recognition, facial attribute prediction, valence-arousal estimation, gender recognition, age prediction, and head pose estimation. See {\citep{jour/TPAMI/Sariyanidi15, jour/TVCG/Tam13, jour/arXiv/Wang14, tech/MSR/Zhang10}} for details.

{\textbf{Registration-based face analysis.}} Despite significant advances in deep learning, automatic face analysis tasks, such as smile detection (a comparative review is provided in Table \ref{table: comparison}), attribute prediction {\citep{conf/AAAI/Hand17, jour/PAMI/Han17, jour/PRL/Sethi18}} and valence-arousal estimation {\citep{jour/TAC/Mollahosseini17}}, still face major challenges caused by occlusions and variances of head pose, scale, and illumination. These challenges are the main reason why every state-of-the-art approach to face analysis requires a pre-normalisation step involving face detection and registration (rotation, scaling, and 2D/3D transformation).

{\textbf{Approaches without pre-normalisation.}} There exist some works that process the original input image without pre-normalisation steps. Liu et al. {\citep{conf/ICCV/Liu15}} combines LNet localising a face and ANet to predict facial attributes. However, they use EdgeBox {\citep{conf/ECCV/Zitnick14}} that proposes a number of candidate windows to determine the final facial region among the multiple predicted positions scattered by LNet. Before feeding the output of LNet to ANet, this process for narrowing the potential face region is performed several times through several LNet stages. This method uses an actual image as input, but the processes inside the architecture operate as a sequential pipeline.

Ranjan et al. {\citep{conf/FG/Ranjan16}} proposed a deep neural network consisting of multiple branches to handle various face-related tasks. The proposed network uses Selective Search {\citep{conf/ICCV/Sande11}} to generate multi-region proposals. Although the proposed network deals with face classification and face-related tasks in different branches, the face classification and multiple face analysis tasks are performed as separate continuous pipelines. In summary, most of the previous works that do not require a pre-normalisation step follow similar mechanisms to {\citep{conf/ICCV/Liu15, conf/FG/Ranjan16}}, which require region proposal steps in the middle of the process. These region proposal steps typically increase the overall processing time.

\noindent {\textbf{Task-specific model design.}} Several recent approaches address the problem of facial attributes prediction \citep{jour/PAMI/Han17, jour/PRL/Sethi18}. Some propose to use successful face-specific feature representations {\citep{conf/ICB/Zhong16}}, modelling class distributions {\citep{conf/CVPRW/Ehrlich16}} and balancing attributes {\citep{conf/ECCV/Rudd16}}, indirect guiding the categorisation of similar features {\citep{conf/CVPR/Wang16, jour/PRL/Sethi18}}, or direct grouping the relevant attributes {\citep{conf/AAAI/Hand17, jour/PAMI/Han17}}. The best performances (more than $90\%$ accuracy) are obtained by specifically designing a model structure that utilises the relations between relevant attributes {\citep{conf/AAAI/Hand17, jour/PAMI/Han17}}. For this purpose, MCNN-AUX {\citep{conf/AAAI/Hand17}} uses implicit and explicit attribute relationships while DMTL {\citep{jour/PAMI/Han17}} relies on attribute correlation and heterogeneity. R-Codean {\citep{jour/PRL/Sethi18}} proposes a new loss function that incorporates both magnitude and direction of image vectors in feature learning and proposes a framework incorporating a patch-based weighting mechanism. By assigning higher weights to relevant patches for each attribute, the method has similar advantages to grouping relevant attributes.

Compared to the previous studies proposing a task-specific model, we propose a generalised architecture that can be used for attribute prediction as well as other face analysis tasks. Unlike the state-of-the-art methods (specifically in face attribute prediction), the proposed Face-SSD uses similar face-size categories associated with each output layer, and incorporates the Hide-and-Seek {\citep{conf/ICCV/Singh17}} data augmentation method which forces the network to explore the entire face area more extensively during training. This size-based categorisation and simple data augmentation strategy enabled us to achieve performance close to the state of the art (more than $90\%$ accuracy in attribute prediction) without customising the network.

{\textbf{Multi-task facial analysis.}}
While most works on facial analysis use a specially designed architecture to tackle a single application, some works attempt to use an integrated single architecture with multiple branches to address multiple tasks. Ranjan et al. {\citep{conf/FG/Ranjan16}} categorised face-related tasks into two groups: subject-independent tasks (e.g., keypoint detection, pose and smile) and subject-dependent tasks (e.g., gender and facial identity). The all-in-one network {\citep{conf/FG/Ranjan16}} learns multiple tasks in a single architecture, but first uses subject-independent class results to register faces. Then the network performs subject-dependent classification tasks sequentially.

Chang et al. {\citep{conf/CVPRW/Chang17}} proposed FATAUVA-Net to learn multiple tasks related to affect behaviours in a single deep neural network. Similarly to the all-in-one network {\citep{conf/FG/Ranjan16}}, FATAUVA-Net categorised similar tasks that share a feature layer. For example, the network branches eye-related tasks (attributes: eyeglasses, narrow eyes / Action Unit (AU): AU6, AU7, AU45) from the same previous layer. The network branches mouth-related tasks (attributes: mouth slightly open, smile / AU: AU23, AU24, AU25) from other layers extracting mouth-related features. The network branches the valence and arousal prediction layer from the associated AU layers. Although these architectures predict multiple face analysis tasks, it is still difficult to generalise or use them in other tasks that take advantage of large patterns for face detection and use small patterns for other face analysis tasks.

{\textbf{Object (face) detection in the wild.}} 
The proposed Face-SSD is inspired by SSD {\citep{conf/ECCV/Liu16}} considering the face as a specific type of object. SSD {\citep{conf/ECCV/Liu16}} has been applied and extended in many research domains, including text detection {\citep{conf/iccv/He17}}, face detection {\citep{conf/ICCV/SZhang17}}, object pose estimation {\citep{conf/3DV/Poirson16, conf/iccv/Kehl17}} and temporal action detection {\citep{conf/MM/Lin17}}. Similar to the latest methods Face-SSD uses the concept of default box {\citep{conf/ECCV/Liu16}} or anchor box {\citep{jour/TPAMI/Ren16}}. 
Using a baseline architecture that has been successfully applied to various detection tasks, we propose the first SSD-inherited architecture that tackles both continuous large-pattern-leveraged tasks (e.g., face detection) and small-pattern-leveraged tasks (e.g., face analysis) in parallel.

\section{The Proposed Framework: Face-SSD}
\label{sec: method}
The proposed Face-SSD framework is shown in Fig. \ref{fig:basic_structure}. Face-SSD is a fully convolutional neural network consisting solely of convolutional and pooling layers. The input to the Face-SSD is a colour image $I$. There are six layers (i.e. $S = 6$), each corresponding to a certain scale, that is size of face. At each scale $s \in [1 \dots 6]$ the output is a heatmap $A$ containing at each spatial position $i$ the confidence score that a face is present at that location, a heatmap $B$ with the parameters of the bounding box of the face associated with that position $i$, and a heatmap $\Gamma$ with the face analysis task confidence score(s) at each position $i$, that is $\{\alpha_{i}, \beta_{i}, \gamma_{i}\}_{s}$ at every spatial location $i$, as shown in Fig. \ref{fig:basic_structure}(b). At test time, a threshold at the face detection confidence score heatmap $A$ selects candidate faces at several spatial locations $i$. Subsequently, Non-Maximum Suppression (NMS) {\citep{conf/ICPR/Neubeck06}} is used to derive the bounding boxes and in each of them calculate scores for the face analysis tasks and for the face detection. 

The following sections describe how to configure Face-SSD (Sec. \ref{model_construction}), how to train face detection and face analysis task in a single architecture (Sec. \ref{subsec: training}), and how to combine the face detection and analysis results during testing (Sec. \ref{subsubsec: testing}).

\subsection{Model Construction}
\label{model_construction}

Face-SSD consists of layers performing at various stages feature extraction (VGG16 Conv. Layers), face detection, and face analysis as shown in Fig. \ref{fig:basic_structure}(a). $G[1:10]$ represents convolution and pooling layer groups with the same input resolution. For example, G2 consists of two convolution layers and one pooling layer, whereas G6 consists of two convolution layers. Similarly to SSD {\citep{conf/ECCV/Liu16}}, Face-SSD outputs six-scale ($S=6$) heatmap volumes generated by multiple output convolution layers [(f1, t1):(f6, t6)]. f[1:6] is produced by the face detection part, while t[1:6] is produced by the face analysis part. The output convolution layers of the two different parts are aligned and concatenated at the end.

Each concatenated output convolution layer outputs a pixel-wise heatmap volume consisting of $(1+4+n)$ heatmap planes. For example, the concatenated output convolution layer for the second scale ($s=2$) outputs a three-dimensional volume ($HM_2 \times HM_2 \times (1+4+n)$) consisting of $(1+4+n)$ heatmap planes having the same resolution ($HM_2 \times HM_2$) of the second scale, as shown in Fig. \ref{fig:basic_structure}(b). The first plane indicates the existence of a face. The next four heatmap planes at each spatial position $i$ contain the centre $(cx, cy) \in R^2$ of the face bounding box and its width $w$, and height $h$. The former is relative to the location $i$ (i.e., $(cx, cy)$ are actually offsets) and the latter is relative to the current heatmap scale $s$. The remaining set of $n$ heatmap planes are the confidences for the $n$ face analysis tasks -- note that these are also heatmaps, that is, they have spatial dimensions as well.

All of the convolution layers are followed by ReLU activation function except for the output convolution layer. For the output convolution layer, for binary classification tasks, such as face classification, smile recognition and attribute prediction, we use the sigmoid function (see Fig. \ref{fig:basic_structure}(b), (b-1) and (b-2), respectively). For regression tasks such as bounding box offsets and valence-arousal estimation, we use linear functions similarly to SSD {\citep{conf/ECCV/Liu16}} (see Fig. \ref{fig:basic_structure}(b) and (b-3)). The parameters for the layers in Face-SSD are summarised in Table \ref{table: network parameter detail}. The parameters of the convolution layer are denoted in the order of number of kernels, kernel size, stride and padding, while the parameters of the pool layer follow the order of kernel size, stride and padding.

During training, the output (prediction) values that appear in heatmaps responsible for the bounding box and tasks are examined only when the corresponding face label exists in the pixel (see details in Sec. \ref{subsubsec: face detection}). During testing, the values for the bounding box and the task-related output are examined only when the corresponding face confidence score exceeds a threshold. The face detection threshold is determined by selecting the optimal value that provides the best performance on the face detection task.

\begin{table}[!t]
\small
\caption{The detailed parameters of Face-SSD layers (see text)}
\label{table: network parameter detail}
\centering
\begin{tabular}{|c|c|c|}
\hline
Group ID			& Conv. ID: Parameters		& Pool	\\ \hline\hline
G1				& [1:2]: (64, 3, 1, 1)			& (2, 2, 0)	\\ \hline
G2				& [1:2]: (128, 3, 1, 1)			& (2, 2, 0)	\\ \hline
G3				& [1:3]: (256, 3, 1, 1)			& (2, 2, 0)	\\ \hline
G4				& [1:3]: (512, 3, 1, 1)			& (2, 2, 0)	\\ \hline
G5				& [1:3]: (512, 3, 1, 1)			& (3, 1, 1)	\\ \hline
\multirow{2}{*}{G6}	& 1: (1024, 3, 1, 1) 			& $\cdot$	\\
 				& 2: (1024, 1, 1, 0) 			& $\cdot$	\\ \hline
\multirow{2}{*}{G7}	& 1: (256, 1, 1, 0)				& $\cdot$	\\
				& 2: (512, 3, 2, 1)				& $\cdot$	\\ \hline
\multirow{2}{*}{G8}	& 1: (128, 1, 1, 0)				& $\cdot$	\\
				& 2: (256, 3, 2, 1)				& $\cdot$	\\ \hline
\multirow{2}{*}{G9}	& 1: (128, 1, 1, 0)				& $\cdot$	\\
				& 2: (256, 3, 1, 0)				& $\cdot$	\\ \hline
\multirow{2}{*}{G10}	& 1: (128, 1, 1, 0)				& $\cdot$	\\
				& 2: (256, 3, 1, 0)				& $\cdot$	\\ \hline
\multirow{3}{*}{Out. Conv.}		& $C_{f}$: (1, 3, 1, 1)		& $\cdot$		\\
				& $B$: (4, 3, 1, 1)				& $\cdot$	\\
				& $C_{t}$: (n, 3, 1, 1)			& $\cdot$	\\
\hline
\end{tabular}
\end{table}

\subsubsection{Implementation details}
\label{subsec:SmileNet_Insignts}

\begin{figure*} [t!]
\centering
      \includegraphics[width=0.95\linewidth]{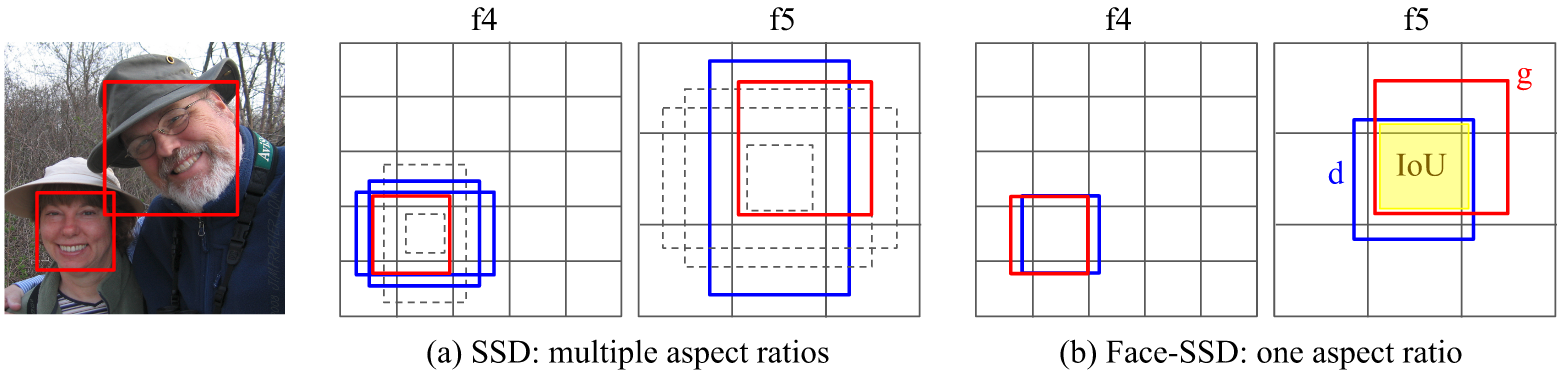}
\caption{ {Example of matched default box for the face confidence heatmaps ${C_f}_{[4:5]}$, produced by $f4$ and $f5$ output convolution layers (see Fig. \ref{fig:basic_structure}). (a) Dotted boxes (grey) represent multiple candidate default boxes with multiple different aspect ratios. Face-SSD (b) uses only one aspect ratio in the matching process of the default box $d$. The example image is one of the sample images of AFLW dataset {\citep{conf/IWBFIAT/Koestinger11}}.}}
\label{fig:default_box_matching_ex}
\end{figure*}

\noindent {\textbf{Single aspect ratio:}} We utilise only one aspect ratio (square) configuring a default box to assign a ground truth label to a pixel position in a heatmap, as shown in Fig. \ref{fig:default_box_matching_ex}. This is because face deformations, caused by expression and pose, result in
similar aspect ratios. This is in accordance with the related work in the literature -- e.g., Hao et al. {\citep{conf/CVPR/Hao17}} proposed Single-Scale RPN utilising one anchor box and Zhang et al. {\citep{conf/ICCV/SZhang17}} proposed S$^{3}$FD utilising one default box.

\noindent {\textbf{Usage of pre-trained models:}} Several works including Liu et al. {\citep{conf/ICCV/Liu15}} demonstrate that models pre-trained on object recognition (e.g., ImageNet {\citep{conf/CVPR/Deng09}}) are useful for face localisation. Similarly, networks pre-trained on face recognition (e.g., CelebFaces {\citep{conf/CVPR/Sun14}}) are useful for capturing face attributes at a more detailed level. For this reason, we selectively use pre-trained parameters (trained with an object dataset {\citep{jour/IJCV/Russakovsky15, conf/ICLR/Simonyan115}} and a face dataset {\citep{conf/IWBFIAT/Koestinger11}}) to initialise the convolution filters for face detection and analysis tasks (see details in Sec. \ref{subsec: training}). This usage of pretrained models helps with improving the Face-SSD performance for both face detection (utilising large patterns) and analysis (utilising relatively smaller patterns) tasks.

\subsection{Training}
\label{subsec: training}

Training of Face-SSD follows the following four steps:
\begin{enumerate}
  \item Copying parameters of the VGG16 network {\citep{conf/ICLR/Simonyan115}} (convolution layers) to the VGG16 (feature extraction) part $G[1:5]$ of Face-SSD and subsampling\footnote{For example, the first fully connected layer $fc6$ of the VGG16 network {\citep{conf/ICLR/Simonyan115}} connects all the positions of a $T_{i} = (f_{vi}, m, m) = (512, 7, 7)$ dimensional input feature map, where $f_{vi}$ is the feature (kernel) dimension at each of the $m^2$ spatial locations, to a $f_{vo} = 4096$ dimension output vector $T_{o}$. Let us organise the weights in a tensor $W_{vgg}$ with dimensions $(f_{vo}, f_{vi}, m, m) = (4096, 512, 7, 7)$. On the other hand, Face-SSD takes an input feature map with dimensions $(512, 18, 18)$ and outputs a feature map with dimensions $T^{\prime}_{o}$ = $(a, m^{\prime}, m^{\prime}) = (1024, 18, 18)$ using filters with  kernel size $3\times3$. The weight tensor $W_{fssd}$ is then of dimensions $(1024, 512, 3, 3)$. In order to initialise the $W_{fssd}$, we uniformly subsample the $W_{vgg}$ along each of its modes -- in our case by a factor $(4,1,3,3)$. This corresponds to subsampling by a factor of $4$ along the dimension of the output feature vector $T_{o}$ and by a factor of $3$ along each spatial dimension of the input tensor $T_{i}$ of the VGG16 network -- we copy the corresponding weights.} the parameters from fully connected layers ($fc6$ and $fc7$) of VGG16 network to the $G6$ layers of Face-SSD, as described in SSD {\citep{conf/ECCV/Liu16}}.
  \item Freezing the face analysis part and finetuning the face detection part by using the AFLW (face) dataset {\citep{conf/IWBFIAT/Koestinger11}}.
  \item Copying the parameters of the layers $G[4:10]$ constituting the face detection part to the corresponding layers of the face analysis part.
  \item Freezing the face detection part and finetuning the layers $G[4:10]$ constituting the face analysis part by using task-related datasets (e.g., CelebA {\citep{conf/ICCV/Liu15}} or GENKI-4K {\citep{GENKI_DB}} for smile recognition, CelebA {\citep{conf/ICCV/Liu15}} for facial attribute prediction, AffectNet {\citep{jour/TAC/Mollahosseini17}} for valence-arousal estimation).
\end{enumerate}

The first and second steps are similar to the initialisation and end-to-end learning process of SSD network {\citep{conf/ECCV/Liu16}}. We use the same cost function as the SSD to finetune the face detection part of Face-SSD. 

\subsubsection{Face Detection}
\label{subsubsec: face detection}

As described above, finetuning of the face detection part is based on the use of an objective loss function $L_{face}$, which is a weighted sum of the face classification loss $L_{cls}$ and the bounding box regression loss $L_{reg}$ defined as:
\begin{equation}
	L_{face}(x_{f}, c, l, g) = \frac{1}{N}( L_{cls}(x_{f}, c) + \lambda x_{f} L_{reg}(l, g) ),
\label{eq:objective_function}
\end{equation}
where N is the total number of matched default boxes. For the regression loss $L_{reg}$, smooth L1 loss {\citep{conf/ICCV/Girshick15}} is used for calculating the distance between the predicted $l=\{l_{cx}, l_{cy}, l_w, l_h\}$ and the ground truth $g=\{g_{cx}, g_{cy}, g_w, g_h\}$ bounding boxes {\citep{conf/ECCV/Liu16}}, as shown in Eq. \ref{eq:reg_loss} and \ref{eq:smooth function}. Specifically,

\begin{equation}
\begin{split}
	L_{reg}(l, g) = \sum\limits_{m \in \{cx,cy,w,h\}} smooth_{L_{1}} (l_m - \hat{g}_m), \\
	\hat{g}_{cx} = (g_{cx} - d_{cx}) / d_w,\indent \hat{g}_{cy} = (g_{cy} - d_{cy}) / d_h, \\
	\hat{g}_{w} = \log(g_{w} / d_{w}),\indent \hat{g}_{h} = \log(g_{h} / d_{h}),
\label{eq:reg_loss}
\end{split}
\end{equation}
where
\begin{equation}
 smooth_{L_{1}} (k) = \begin{cases}
 				0.5 k^{2},					& \text{if } \| k \| < 1 \\
				\| k \| - 0.5,				& \text{otherwise}
 				\end{cases}
\label{eq:smooth function}
\end{equation}

The face classification loss $L_{cls}$ is based on binary cross entropy over face confidence scores $c$, as shown in Eq. \ref{eq:cls loss}.

\begin{equation}
	L_{cls}(x_{f},c) = -x_{f} log(c) - (1 - x_{f}) log (1 - c )
\label{eq:cls loss}
\end{equation}

The flag $x_{f}\in\{1,0\}$, used in the equations above is set to 1 when the overlap between the ground truth and the default bounding box $d=\{d_{cx}, d_{cy}, d_w, d_h\}$ exceeds a threshold. Note that the regression loss is only used when $x_{f}=1$, and is disabled otherwise. 

At the later stages of the training, similar to {\citep{conf/ECCV/Liu16}} we use Hard Negative Mining (HNM), that is, we sort calculated losses only in the background region ($\neg(x_{f}=1)$) in descending order and select and backpropagate only from the highest ones. Following {\citep{conf/ECCV/Liu16}}, we set the loss-balancing weight $\lambda$ (in Eq. \ref{eq:objective_function}) to $1$.

\subsubsection{Face Analysis}
\label{subsubsec: task analysis}

This section describes how to apply Face-SSD to various face analysis tasks. We address three problems: smile recognition as binary classification, facial attribute prediction as multi-class recognition and valence-arousal estimation as multi-task regression. In all three problems, the architecture of the network differs only in terms of the number $n$ of the facial task heatmaps.
For datasets that have multiple annotations for the same image, Face-SSD supports multi-task learning by defining a multi-task loss function as in Eq. \ref{eq:task loss}.

\begin{equation}
	L_{total} = \sum\limits_{t = 1}^{T} ||w_{t} L_{t}(g_{t}, p_{t})||_{2}, 
\label{eq:task loss}
\end{equation}

That is, the multi-task loss $L_{total}$ is defined as the $L2$ norm of multiple weighted individual face analysis task losses $\{w_{t} L_{t}\}$. $L_{t}$ is used to calculate errors using a ground truth $g_{t}$ and a prediction $p_{t}$ for a given task $t$. $T$ denotes the total number of face analysis tasks. In what follows we define the loss functions used for different problems we address.

\vspace{5mm} 

\label{subsubsec: smile recognition}
\noindent {\textbf{Smile Recognition.}} 
The smile classification loss $L_{smile}$, is the binary cross entropy over smile confidence scores $e$ and the ground truth $x_{e}=\{1,0\}$ as defined in Eq. \ref{eq:emotion_loss}. 

\begin{equation}
	L_{smile}(x_{e}, e) = -x_{e} log(e) - (1 - x_{e}) log (1 - e )
\label{eq:emotion_loss}
\end{equation}

The ground truth $x_{e}=\{1,0\}$ at each location is set using the default box matching strategy {\citep{conf/ECCV/Liu16}}. The loss is defined at each spatial location of the output heatmap, and in this case, we do not use Hard Negative Mining (HNM), which was required to select negative samples for face detection (see Sec. \ref{subsubsec: face detection}).

Finetuning the network for face analysis tasks (i.e., smile recognition) does not impair the face detection performance due to freezing the parameters for the face detection part of Face-SSD.

\vspace{5mm} 

\label{subsubsec: multi-attribute learning}
\noindent {\textbf{Facial Attribute Prediction.}} Facial attribute prediction is treated as multiple binary classification problems where a number of attributes may exist simultaneously. For example, a face attribute (such as smiling) can appear independently of other attributes (such as the gender or hair colour). Therefore, we define the facial attribute prediction loss $L_{att}$ as the average of independent attribute losses, that is 
\begin{equation}
	L_{att} (G, P) = - \frac{1}{N_a} \sum\limits_{a = 1}^{N_a} (g_{a} log(p_{a}) + (1 - g_{a}) log (1 - p_{a})),
\label{eq:att_loss}
\end{equation}
where $N_a$ denotes the total number of attributes. $g_{a} \in G$ and $p_{a} \in P$ denote the ground truth (1 or 0) label and a predicted attribute confidence score of the $a$-th attribute, respectively. For calculating a single attribute prediction loss associated with an individual attribute $a$, we use the binary cross entropy over attribute confidence scores $p_{a}$.

\vspace{5mm} 

\label{subsubsec: v-a estimation}
\noindent {\textbf{Valence and Arousal Estimation.}} 
Similar to several other previous works (e.g. 
{\citep{jour/IVC/Koelstra13}}, {\citep{jour/TAC/Mollahosseini17}}), we treat arousal and valence prediction as a regression problem. Valence is related to the degree of positiveness of the affective state, whereas arousal is related to the degree of excitement {\citep{jour/PR/Russell03, jour/JPSP/Russell99}}. We used the Euclidean (L2) distance between the predicted value $\hat{y}_n$ and ground truth value of valence/arousal $y_{n}$, as shown in Eq. \ref{eq:euclidean_dist}. The loss is then defined as the sum of the valence $E_v$ and the arousal $E_a$ losses, that is
\begin{equation}    
\begin{split}    
	L_{emo} = E_v + E_a, \\
    E = \frac{1}{2N} \sum\limits_{n=1}^{N} || \hat{y}_n - y_{n} ||_{2}^{2},
\label{eq:euclidean_dist}
\end{split}
\end{equation}
where $N$ is the number of image samples in a mini-batch. 

\subsubsection{Data Augmentation in Training}
\label{subsec:data_augmentation}

Face-SSD uses a $300 \times 300$ resolution and $3$ channel colour input image. Prior to data augmentation, all pixel values for the R, G, and B channels of a sample image are normalised based on the mean and standard deviation values of the entire dataset. Each sample image is first flipped in the horizontal direction with a probability of 0.5. In the training session, we randomly select one of data augmentation mechanisms (shrinking, cropping, gamma correction and Hide-and-Seek (H-a-S) {\citep{conf/ICCV/Singh17}}) to create noisy data samples for each epoch.

Both shrinking and cropping maintain the aspect ratio. Gamma correction is applied separately to the individual R, G, B channels. In Hide and Seek (H-a-S) {\citep{conf/ICCV/Singh17}} we hide image subareas and force a network to seek more context in areas that are not as discriminative as key distinctive areas such as lip corners. We first randomly select a division number among $3$, $4$, $5$ or $6$. If we select $3$, the image region will be divided into $9$ ($3 \times 3$) sub-image patches. Each sub-image patch is then hidden (filled with the mean R, G, B values of all data samples in a dataset) with a probability of $0.25$.

\subsection{Testing}
\label{subsubsec: testing}

\begin{figure} [t!]
\centering
    \includegraphics[width=0.95\linewidth]{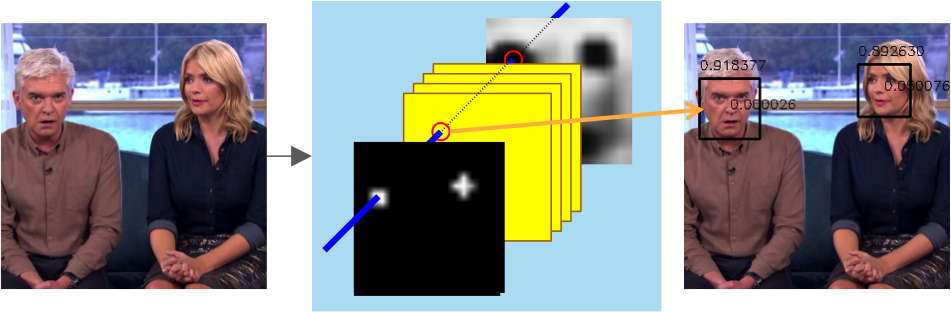} 
    \subfigure[Smile Recognition]
    {
    	\label{subfig:smile_results_ex}
	\includegraphics[width=0.95\linewidth]{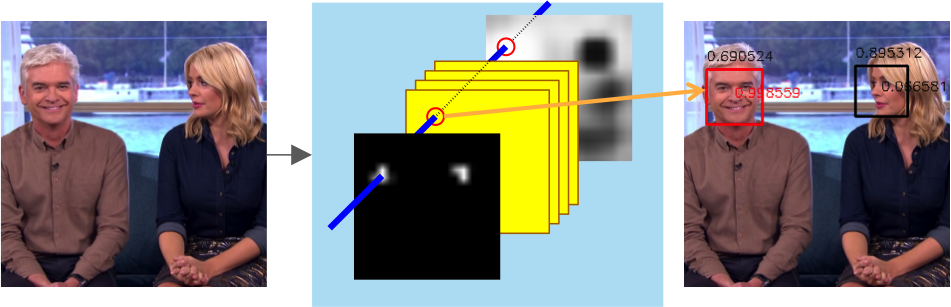}
    }
    \subfigure[Valence-Arousal Estimation]
    {
         \label{subfig:v_a_results_ex}
	\includegraphics[width=0.95\linewidth]{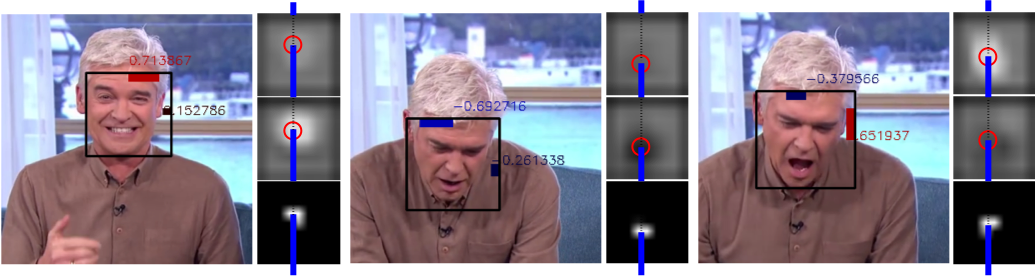}
    }
\caption{{Examples of face detection and face analysis tasks. As a representative example of classification and regression, we visualised the output heatmaps for smile recognition and valence-arousal estimation. \subref{subfig:smile_results_ex} The heatmaps represent face classification, bounding box regression and smile recognition results. \subref{subfig:v_a_results_ex} For the valence-arousal example, we only visualise the output heatmaps for face classification, valence and arousal estimation from the bottom row. We rescaled the range of output values at the valence-arousal estimation heatmap from $[-1:1]$ to $[0:255]$ for the visualisation. The median (127) in this example represents the neutral valence or arousal value (0).}}
\label{fig:ex_face_emotion_detection}
\end{figure}

The registration-free Face-SSD for a specific face analysis task (e.g., smile recognition) is based on both face and task (e.g., smile) confidence scores. First, the locations in the face confidence heatmap, for which the score exceeds a threshold ($th_{face} = 0.1$), are selected. Then Non-Maximum Suppression (NMS) method (with jaccard overlap value $0.35$ as in S$^{3}$FD {\citep{conf/ICCV/SZhang17}}) is used to extract the final bounding boxes. Subsequently, a task-specific threshold $th_{t}$ is applied on the task related score of the final bounding boxes (Fig. \ref{fig:ex_face_emotion_detection}). In the case of the regression (e.g., valence-arousal estimation), the output value of the final bounding box is used.

As mentioned in Sec. \ref{model_construction}, each output layer of Face-SSD generates several heatmaps: one for face detection, four for the offset coordinates of face bounding box and $n$ for the $n$ number of face analysis tasks, as shown in Fig. \ref{fig:basic_structure}(b). Specifically, Fig. {\ref{subfig:smile_results_ex}} and {\subref{subfig:v_a_results_ex}} visualise the heatmaps generated by Face-SSD's second and third-scale output layers ($s=2, 3$), which handle the second and third smallest sizes of the face that appears in the image, respectively. Thus, activations in the heatmap are high when a specific size of face is detected. For the given example of smile recognition, as shown in Fig. {\ref{subfig:smile_results_ex}}, the forefront heatmap shows two clusters of pixels, indicating the existence of two faces. The rearmost heatmap highlights the corresponding pixel only when a task is detected. In this example the heatmap has high values when the detected face is a smiling face.

\section{Experiments and Results}
\label{sec: experiments}
	
\subsection{Datasets}
\label{subsec: exp datasets}

In this paper, we show the performance of the proposed Face-SSD on three representative face analysis applications such as smile recognition (binary classification), facial attribute prediction (multiple class recognition), and valence-arousal estimation (multiple task regression). We stress that the structure of the network, including the number of filters and filter sizes remain the same -- the only change is the number of output layer heatmaps. We used GENKI-4K {\citep{GENKI_DB}}, CelebA {\citep{conf/ICCV/Liu15}}, and AffectNet {\citep{jour/TAC/Mollahosseini17}} datasets to test the three representative applications using Face-SSD.

Beginning with {\citep{jour/TPAMI/Whitehill09}}, which performed the first extensive smile detection study, most of the subsequent studies used the GENKI-4K\footnote{The GENKI-4K {\citep{GENKI_DB}} dataset is a subset of the GENKI dataset used in {\citep{jour/TPAMI/Whitehill09}}. This dataset consists of $4,000$ face images, each labelled with smile and head pose (yaw, pitch, roll). Only the GENKI-4K dataset is publicly available.} dataset for performance evaluation {\citep{GENKI_DB}}. In this paper, the smiling face detection experiments were performed not only on the GENKI-4K dataset but also on the CelebA dataset {\citep{conf/ICCV/Liu15}} which also contains smile labels. For facial attribute prediction experiments, we used the CelebA dataset {\citep{conf/ICCV/Liu15}} which is the most representative dataset. Finally, for the valence-arousal estimation experiment we used the AffectNet {\citep{jour/TAC/Mollahosseini17}} dataset consisting of continuous level (valence-arousal) labels and face images captured in the wild.

The AFLW dataset {\citep{conf/IWBFIAT/Koestinger11}} used for face detection and other datasets used for face analysis tasks (i.e., GENKI-4K {\citep{GENKI_DB}}, CelebA {\citep{conf/ICCV/Liu15}}, AffectNet {\citep{jour/TAC/Mollahosseini17}}) have different bounding box positions and shapes. To solve this problem, we empirically adjusted the bounding box position of these datasets to create a square box that surrounds the entire face area centred on the nose (similar to the bounding box of the AFLW dataset). To do this, we first used the trained Face-SSD to detect a face bounding box. Then, we double-checked whether the detected bounding box is correct. If it was incorrect, we modified the bounding box manually.

\begin{figure} [t!]
\small
\begin{center}
      \includegraphics[width=0.95\linewidth]{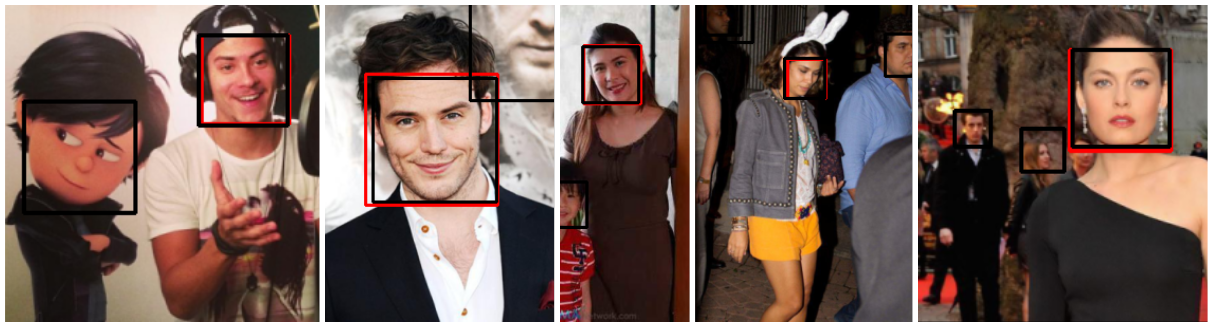}
\end{center}
  \vspace{-0.4cm}
\caption{{If there were multiple faces detected (black boxes), only the annotated faces with the ground truth label ({\color{R}{red box}}) were evaluated for attribute prediction. The face detected in the background was not used for accuracy measurement.}}
\label{fig:multi-detection case for attribute testing}
\end{figure}

In particular, when using the CelebA {\citep{conf/ICCV/Liu15}} dataset, we only examined smile recognition and facial attribute prediction performance for annotated faces. Each image sample in the CelebA dataset has only one bounding box with its corresponding attribute labels, even if the image contains multiple faces. Therefore, when multiple bounding boxes were detected (black boxes in Fig. \ref{fig:multi-detection case for attribute testing}) during the test time, we only calculated the accuracy for the detected bounding box that matched the ground truth position (red box in Fig. \ref{fig:multi-detection case for attribute testing}). If there is no bounding box detected for the ground truth location, it is considered as a false negative when calculating the accuracy.

\subsection{Face Detection}
\label{subsec: face detection performance}

First, we evaluate the face detection performance. Although Face-SSD performs face detection in parallel with one or more tasks, the face analysis task results appearing in the output heatmap are only examined at the corresponding pixel positions that indicate successful face detection (as discussed in Sec. \ref{subsubsec: testing}).

Here, we evaluate the face detection performance of Face-SSD on face analysis task datasets, including GENKI-4K {\citep{GENKI_DB}}, CelebA {\citep{conf/ICCV/Liu15}}, and AffectNet {\citep{jour/TAC/Mollahosseini17}}. According to {\citep{jour/JoV/Du11}}'s experimental results, the visual recognition ability of a human is degraded when image resolution falls below $20 \times 30$ pixels. For this reason, the face detection of Face-SSD aims to support face analysis tasks rather than detecting tiny faces, which is beyond the scope of this work. To this end, we evaluate the face detection performance on the face analysis task (e.g., smile, attribute, valence-arousal) datasets that do not include severe occlusion or very small faces. Instead, these datasets consist of images that typically contain high-resolution faces compared to $20 \times 30$ pixels and are captured in the wild (with naturalistic variations in pose, occlusion, and/or scale).

The face detection results are shown in Table \ref{table: FD_HNM_HaS_Effects} in terms of Equal Error Rate (EER) and Average Precision (AP) {\citep{jour/IJCV/Everingham10}}. First, we investigated face detection performance using the same strategy as the SSD {\citep{conf/ECCV/Liu16}} called Face-SSD Baseline (Face-SSD-B) {\citep{conf/ICCVW/Jang17}}. The AFLW dataset {\citep{conf/IWBFIAT/Koestinger11}} was used for training face detection part of Face-SSD. For data augmentation, Face-SSD-B used shrinking, cropping, and gamma correction (see details in Sec. \ref{subsec:data_augmentation}). Using the data augmentation, Face-SSD-B trained on the non-challenging face dataset AFLW did not achieve a competitive performance (EER=$05.42\%$ and AP=$99.50$) in comparison to using other face detection datasets. However, unlike general face detection evaluation, we used the simplest face analysis task dataset (GENKI-4K {\citep{GENKI_DB}}) to provide a performance comparison between different strategy combinations.

\begin{table*}[!t]
\small
\caption{Effects of using Hard Negative Mining (HNM) and Hide-and-Seek (H-a-S) methods when training face detection in Face-SSD. (See text for more details about abbreviations and description)}
\label{table: FD_HNM_HaS_Effects}
\centering
\begin{tabular}{|c||c c|c|c c|c c||c|c|}
\hline
\multirow{2}{*}{ }		& \multicolumn{2}{ |c| }{IoU for GTs}	
& \multirow{2}{*}{HNM}	& \multicolumn{2}{ |c| }{H-a-S for All}	
& \multicolumn{2}{ |c|| }{H-a-S for Half}	
& \multicolumn{2}{ |c| }{GENKI-4K Test Results} \\
\cline{2-3}\cline{5-10}
						& 0.50 			& 0.35
&  	 					& Fine 	 		& Coarse
& Fine 					& Coarse
& EER ($\%$) 			& AP \\
\hline\hline
{\textbf{Face-SSD-B}}aseline {\citep{conf/ICCVW/Jang17}}	& $\checkmark$ 	& $\cdot$	
& $\cdot$				& $\cdot$ 		& $\cdot$	
& $\cdot$ 				& $\cdot$			
& 05.42					& 99.50 \\
\hline
Face-SSD-B with More GTs 	& $\cdot$ 		& $\checkmark$	
& $\cdot$				& $\cdot$ 		& $\cdot$	
& $\cdot$ 				& $\cdot$			
& {\textbf{03.68}}		& {\textbf{99.91}} \\
\hline
Face-SSD-B with HNM		& $\cdot$ 		& $\checkmark$	
& $\checkmark$			& $\cdot$ 		& $\cdot$	
& $\cdot$ 				& $\cdot$			
& {\textbf{01.72}}		& {\textbf{99.88}} \\
\hline
\multirow{4}{*}{Face-SSD-B with H-a-S} & $\cdot$ & $\checkmark$	
& $\cdot$				& $\checkmark$ 	& $\cdot$ 
& $\cdot$ 				& $\cdot$	
& 34.83					& 93.54 \\
						& $\cdot$ 		& $\checkmark$	
& $\cdot$ 				& $\cdot$ 		& $\checkmark$ 
& $\cdot$ 				& $\cdot$	
& 08.26					& 97.79 \\
						& $\cdot$ 		& $\checkmark$	
& $\cdot$				& $\cdot$ 		& $\cdot$ 
& $\checkmark$ 			& $\cdot$	
& 01.95					& 99.89 \\
						& $\cdot$ 		& $\checkmark$	
& $\cdot$				& $\cdot$ 		& $\cdot$ 
& $\cdot$ 				& $\checkmark$	
& {\textbf{01.16}}		& {\textbf{99.91}} \\
\hline
{\textbf{Face-SSD}} & $\cdot$ & $\checkmark$
& $\checkmark$			& $\cdot$ 		& $\cdot$ 
& $\cdot$ 				& $\checkmark$	
& {\textbf{00.66}}		& {\textbf{99.88}} \\
\hline
\end{tabular}
\end{table*}

To improve the face detection performance we first lowered the IoU threshold from $0.50$ to $0.35$ when assigning ground truths, similarly to S$^{3}$FD {\citep{conf/ICCV/SZhang17}}. Lowering the IoU threshold when matching default box increases the number of positive examples. By doing so the accuracy was improved from EER=$05.42\%$ and AP=$99.50$ to EER=$03.68\%$ and AP=$99.91$.

In order to improve the performance further, we applied a Hard Negative Mining (HNM) strategy on the training data samples in a minibatch. Specifically, we extracted $30\%$ of the data samples that currently output the largest loss in a minibatch, and then re-used the data samples in the next minibatch. By doing so, we further reduced the detection error from EER=$03.68\%$ and AP=$99.91$ to EER=$01.72\%$ and AP=$99.88$.

Finally, we applied H-a-S {\citep{conf/ICCV/Singh17}} as one of our data augmentation strategies. However, unlike what is reported in the original H-a-S paper {\citep{conf/ICCV/Singh17}}, when the H-a-S method was applied to all training samples, the detection performance dropped significantly to EER=$34.83\%$ and AP=$93.54$. Applying the H-a-S method randomly to approximately half of the training samples reduced the error to EER=$01.95\%$ and AP=$99.89$. In addition, as shown in Table \ref{table: FD_HNM_HaS_Effects}, our results indicate that for face detection it is better to hide coarsely divided patches (EER=$01.16\%$ and AP=$99.91$) than to hide finely divided ones (EER=$01.95\%$ and AP=$99.89$) because face detection relies on relatively large continuous patterns. In Table \ref{table: FD_HNM_HaS_Effects}, for H-a-S, the coarse patch division process randomly selects the patch size from 3, 4, 5 and 6 (see Sec. \ref{subsec:data_augmentation}), whereas the fine patch division process randomly selects the patch splitting size from 16, 32, 44 and 56 as proposed originally in {\citep{conf/ICCV/Singh17}}.

\begin{figure} [t!]
\begin{center}
      \includegraphics[width=0.95\linewidth]{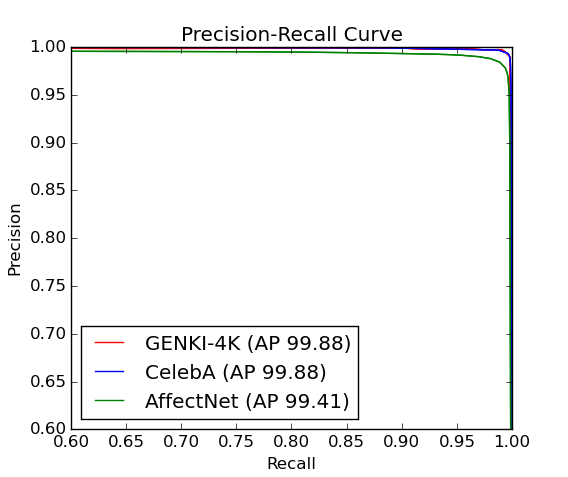}
      \includegraphics[width=0.95\linewidth]{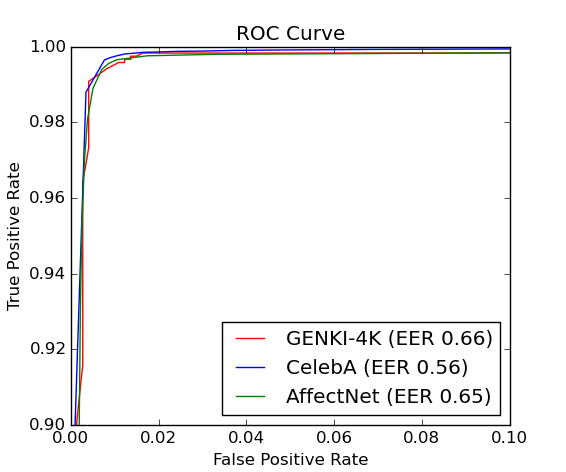}
\end{center}
  \vspace{-0.4cm}
\caption{{Experimental curves for face detection performance on GENKI-4K {\citep{GENKI_DB}}, CelebA {\citep{conf/ICCV/Liu15}} and AffectNet {\citep{jour/TAC/Mollahosseini17}} datasets: Precision-Recall curves and Receiver Operating Characteristic (ROC) curves.}}
\label{fig:exp_face_detection_curves}
\end{figure}

By applying the training strategies of low IoU threshold, HNM and H-a-S, we achieved EER=$0.66\%$ and AP=$99.88$ on the GENKI-4K dataset. For the CelebA dataset, we achieved EER=$0.56\%$ and AP=$99.88$, as shown in Fig. \ref{fig:exp_face_detection_curves}. For the AffectNet dataset, we achieved EER=$0.65\%$ and AP=$99.41$. These results indicate that Face-SSD can robustly detect faces in unconstrained environments, and the Face-SSD can be used for further face analysis tasks such as facial attribute and affect prediction along the dimensions of valence and arousal. The optimal thresholds for the best face detection accuracy were $0.20$ for the GENKI-4K dataset, $0.16$ for CelebA dataset, and $0.11$ for AffectNet dataset.

\subsection{Face Analysis}
\label{subsec: smile recognition performance}


Face-SSD is inspired by SSD {\citep{conf/ECCV/Liu16}}, which promises real-time detection performance. Thus, the parameter values used in the process of finetuning the face detection and the face analysis parts of Face-SSD are initialised with the values used for training the base network of SSD {\citep{conf/ECCV/Liu16}}. We used SGD with initial learning rate=$10^{-3}$, momentum=$0.9$, weight decay=$0.0005$, and batch size=$16$. We used learning rate=$10^{-3}$ for the first $40K$ iterations, then continued training for $40K$ with learning rate=$10^{-2}$. We continuously reduced the learning rate every $40K$ iterations until it reached learning rate=$10^{-5}$. Increasing the learning rate for the second $40K$ iterations speeds up the optimisation process. However, we first started the training process with learning rate=$10^{-3}$, because the optimisation process tends to diverge if we use a larger learning rate in the beginning.

The following sections detail the experiments we have conducted to evaluate the two main performance factors of Face-SSD, namely prediction accuracy and processing time, for two tasks: smile recognition and facial attribute prediction.

\begin{table*}[!t]
\small
\caption{A detailed comparison with the state-of-the-art methods on the GENKI-4K dataset {\citep{GENKI_DB}}. We summarise the features, classifiers, detection / registration methods and input image resolution (width, height, and channel) that were used in previous studies in published order. All previous studies require a normalised (cropped and aligned) input image, which necessarily require face detection and registration steps in advance (except {\citep{jour/MVA/Chen17}}-II and III). Some works {\citep{jour/TIP/Shan12, conf/ICSPRA/Jain13, conf/ACPR/Zhang15, conf/ICIP/Li16, jour/MVA/Chen17}} do not specify how to detect and align a face (in this case, `?'), while {\citep{conf/ECCVW/Kahou14}} mentions that the original image is used if the face detection fails.}
\label{table: comparison}
\centering
\begin{tabular}{|c|c|c|c|c|c|l|}
\hline
Method 						& Feature				& Classifier		& Detection 		& Registration	& Input ($W \times H \times C$)	& Accuracy $(\%)$ \\
\hline\hline
{\citep{jour/TIP/Shan12}}		& Pixel comparison		& AdaBoost		& ? 			& Eyes (manual)		& $48\times48\times1$ 		& $89.70 \pm 0.45$ \\
{\citep{conf/ACCV/Liu12}}		& HOG					& SVM			& VJ* & Eyes	& $48\times48\times1$		& $92.26 \pm 0.81$ \\
{\citep{conf/ICSPRA/Jain13}}	& Multi-Gaussian			& SVM			& VJ* &?	& $64\times64\times1$		& $92.97$ \\
{\citep{conf/ECCVW/Kahou14}}	& LBP					& SVM			& VJ*+Sun* / ori. 			& $5+6$ Pts		& $96\times96\times1$ 		& $93.20 \pm 0.92$ \\
{\citep{jour/NeuroCom/An15}} 	& HOG					& ELM			& VJ*	& Flow-based*	& $100\times100\times1$ 		& $88.20$ \\
{\citep{conf/ACPR/Zhang15}}	& CNN		& Softmax		& ? 			& Face Pts		& $90\times90\times1$ 		& $94.60 \pm 0.29$ \\
{\citep{conf/ICIP/Li16}}		& Gabor-HOG				& SVM			& VJ* / manual 	& ?		& $64\times64\times1$	& $91.60 \pm 0.89$ \\
{\citep{jour/MVA/Chen17}}-I 	& CNN		& SVM			& Liu*			& ?		& $64\times64\times1$ 		& $92.05 \pm 0.74$ \\
\hline
{\citep{jour/MVA/Chen17}}-II 	& CNN		& SVM			& Liu*			& $\cdot$		& $64\times64\times1$ 		& ${\textbf{90.60}} \pm 0.75$ \\
{\citep{jour/MVA/Chen17}}-III	& CNN		& SVM			& $\cdot$ 	& $\cdot$		& $64\times64\times1$ 		& $78.10 \pm 0.56$ \\
\hline
{\textbf{Face-SSD}} 				& {\textbf{CNN}}		& {\textbf{Sigmoid	}}	& $\cdot$ 	& $\cdot$		& ${\textbf{300}}\times{\textbf{300}}\times{\textbf{3}}$ 		& ${\textbf{95.76}} \pm 0.56$ \\
\hline
\end{tabular}
\\ * VJ: {\citep{jour/IJCV/Viola04}}, Liu: {\citep{conf/ICCV/Liu15}}, Sun: {\citep{conf/CVPR/Sun13}}, Flow-based: {\citep{jour/NeuroCom/An15}}
\end{table*}

\subsubsection{Smile Recognition}
\label{subsec: quantitative results}

Accuracy for this task refers to the smile recognition performance including the face detection results. If face detection fails, the result of smile recognition is considered to be a non-smile. 

\begin{figure} [t!]
\small
\begin{center}
      \includegraphics[width=0.95\linewidth]{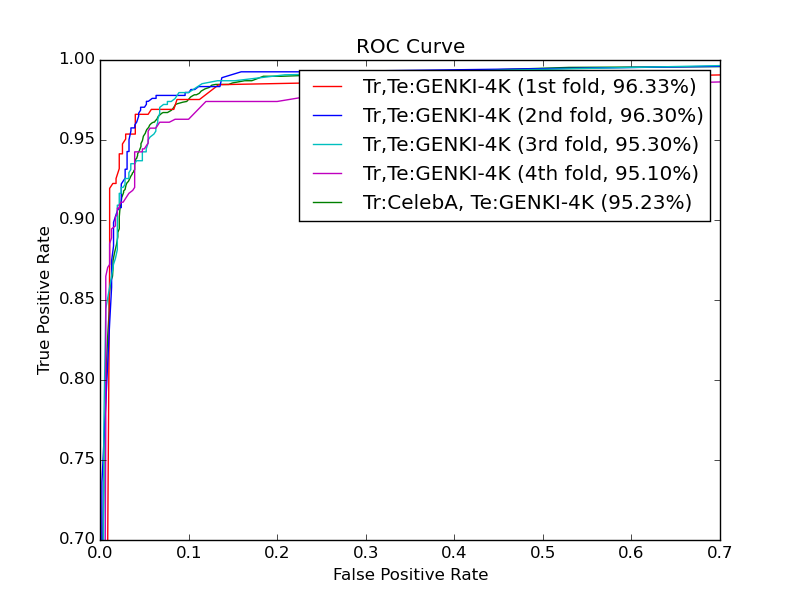}
\end{center}
  \vspace{-0.4cm}
\caption{{Receiver Operating Characteristic (ROC) curve for smiling face detection accuracy using GENKI-4K {\citep{GENKI_DB}} dataset. Tr and Te represent training and testing, respectively.}}
\label{fig:ROC_graph_smile_GENKI}
\end{figure}

{\textbf{Testing on the GENKI-4K dataset:}} Experiments that use this dataset are conventionally based on four-fold validation procedures. However, as GENKI-4K dataset contains a relatively small number of data samples ($4,000$), for training we initially utilised the CelebA dataset that contains a rich set of images. When Face-SSD was trained on the CelebA dataset, we used the entire GENKI-4K dataset for testing. We obtained a smile recognition accuracy of $95.23\%$, as shown in Fig. \ref{fig:ROC_graph_smile_GENKI}. Despite being trained on a completely different dataset with different characteristics, Face-SSD has already surpassed all the latest methods that used the GENKI-4K dataset for testing, as shown in Table \ref{table: comparison}.

To provide a fair comparison with other methods that use the four-fold validation strategy, we used the GENKI-4K dataset together with the bounding box annotations obtained with our method (as explained in Sec. \ref{subsec: exp datasets}) to finetune the Face-SSD, which was trained on the CelebA dataset. In this case, the smile recognition accuracy is improved further. This is due to the fact that the training samples in GENKI-4K dataset are relatively similar to the testing samples as compared to CelebA dataset. Although the training and testing samples do not overlap, using the same dataset (GENKI-4K) for training helps Face-SSD learn the test sample characteristics of the same (GENKI-4K) dataset. Our four-fold validation results were $96.33\%$, $96.30\%$, $95.30\%$ and $95.10\%$, as shown in Fig. \ref{fig:ROC_graph_smile_GENKI}. Compared to the accuracies reported by existing works listed in Table \ref{table: comparison}, our method obtains the best results with mean=${\textbf{95.76}}\%$ and standard deviation=$0.56\%$.

Although Face-SSD does not require separate steps for face detection and registration, Face-SSD's smile recognition results rely on the face detection performed in parallel on the same architecture (as explained in Sec. \ref{subsec: face detection performance}). Among the existing works listed in Table \ref{table: comparison}, Chen's work ({\citep{jour/MVA/Chen17}}-II) reports testing accuracy when the registration process is not used. We therefore compare  Face-SSD's  smile recognition performance more closely to the method of Chen ({\citep{jour/MVA/Chen17}}-II). Our experimental results show that Face-SSD outperforms (${\textbf{95.76}}\%$) the most recently reported smile recognition result of Chen ($90.60\%$) based on a deep learning architecture ({\citep{jour/MVA/Chen17}}-II).

\begin{figure*} [!t]
\begin{center}
      \includegraphics[width=0.95\linewidth]{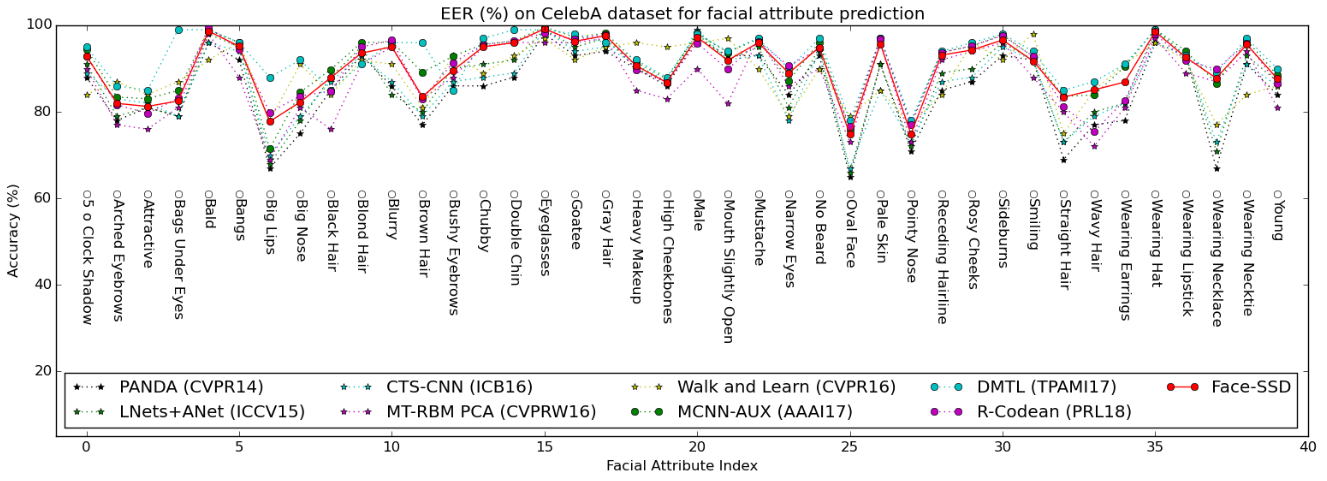}
\end{center}
  \vspace{-0.4cm}
\caption{{Performance comparison in terms of accuracy ($\%$) on CelebA {\citep{conf/ICCV/Liu15}} dataset for facial attribute prediction. Face-SSD delivers excellent prediction performance that is very close to the state-of-the-art models without modifying the Face-SSD architecture. The state-of-the-art models are PANDA {\citep{conf/CVPR/Zhang14}}, LNets+ANet {\citep{conf/ICCV/Liu15}}, CTS-CNN {\citep{conf/ICB/Zhong16}}, MT-RBM PCA {\citep{conf/CVPRW/Ehrlich16}}, Walk and Learn {\citep{conf/CVPR/Wang16}}, MCNN-AUUX {\citep{conf/AAAI/Hand17}}, DMTL {\citep{jour/PAMI/Han17}}, R-Codean {\citep{jour/PRL/Sethi18}}. (See Table \ref{table: comparison_att_CelebA} for more detailed accuracy comparisons.)}}
\label{fig:ACC_multi_att}
\end{figure*}

{\textbf{Testing on the CelebA dataset:}} In the second experiment, we used the CelebA dataset to train and test Face-SSD. In this experiment, we randomly selected $75\%$ of the dataset for training and used the remaining $25\%$ for the testing. We performed several experiments using different combinations of randomly selected training and test samples. Our experimental results show that Face-SSD detects smiling faces accurately (mean=${\textbf{92.81}}\%$), similarly to the state-of-the-art methods ({\citep{conf/ICCV/Liu15}}: $92.00\%$ and {\citep{conf/FG/Ranjan16}}: $93.00\%$), as shown in Table \ref{table: comparison_CelebA}. However, Face-SSD is much faster (${\textbf{47.28}}$ $ms$) than the other methods ({\citep{conf/ICCV/Liu15}}: $139$ $ms$, {\citep{conf/FG/Ranjan16}}: $3,500$ $ms$) that require region proposal methods for smile recognition (see Table \ref{table: comparison_CelebA}).

\begin{table}[!t]
\small
\caption{Comparison to the state-of-the-art methods on the CelebA dataset in terms of accuracy $(\%)$ and time (ms.). RP, EB and SS refer to Region Proposal, EdgeBox {\citep{conf/ECCV/Zitnick14}} and Selective Search {\citep{conf/ICCV/Sande11}}, respectively.}
\label{table: comparison_CelebA}
\centering
\begin{tabular}{|c|c|c|c|}
\hline
Method 									& RP										& Acc. $(\%)$	& Time (ms.)	  \\
\hline\hline
Liu et al. {\citep{conf/ICCV/Liu15}}		& EB {\citep{conf/ECCV/Zitnick14}}			& $92.00$ 					& $139.00$ \\
Ranjan et al. {\citep{conf/FG/Ranjan16}}	&SS {\citep{conf/ICCV/Sande11}}	& $93.00$					& $3,500.00$ \\
{\textbf{Face-SSD}} 							& $\cdot$										& ${\textbf{92.81}}$					& ${\textbf{47.28}}$ \\
\hline
\end{tabular}
\end{table}

\begin{table*}[!t]
\tiny
\caption{Comparison to the state-of-the-art methods for facial attribute prediction on the CelebA dataset in terms of prediction accuracy. The average accuracies of PANDA {\citep{conf/CVPR/Zhang14}}, LNets+ANet {\citep{conf/ICCV/Liu15}}, CTS-CNN {\citep{conf/ICB/Zhong16}}, MT-RBM PCA {\citep{conf/CVPRW/Ehrlich16}}, Walk and Learn {\citep{conf/CVPR/Wang16}}, MCNN-AUUX {\citep{conf/AAAI/Hand17}}, DMTL {\citep{jour/PAMI/Han17}}, R-Codean {\citep{jour/PRL/Sethi18}}, and the proposed Face-SSD are $85.42\%$, $87.30\%$, $86.60\%$, $86.97\%$, $88.65\%$, $91.29\%$, $92.60\%$, $90.14\%$ and $90.14\%$, respectively.}
\label{table: comparison_att_CelebA}
\centering
\begin{tabular}{|c|c|c|c|c|c|c|c|c|c|c|c|c|c|c|c|c|c|c|c|c||c|}
\hline
 &  {\rotatebox[origin=c]{90}{5 o Clock Shadow}}  &  {\rotatebox[origin=c]{90}{Arched Eyebrows}}  &  {\rotatebox[origin=c]{90}{Attractive}}  &  {\rotatebox[origin=c]{90}{Bags Under Eyes}}  &  {\rotatebox[origin=c]{90}{Bald}}  &  {\rotatebox[origin=c]{90}{Bangs}}  &  {\rotatebox[origin=c]{90}{Big Lips}}  &  {\rotatebox[origin=c]{90}{Big Nose}}  &  {\rotatebox[origin=c]{90}{Black Hair}}  &  {\rotatebox[origin=c]{90}{Blond Hair}}  &  {\rotatebox[origin=c]{90}{Blurry}}  &  {\rotatebox[origin=c]{90}{Brown Hair}}  &  {\rotatebox[origin=c]{90}{Bushy Eyebrows}}  &  {\rotatebox[origin=c]{90}{Chubby}}  &  {\rotatebox[origin=c]{90}{Double Chin}}  &  {\rotatebox[origin=c]{90}{Eyeglasses}}  &  {\rotatebox[origin=c]{90}{Goatee}}  &  {\rotatebox[origin=c]{90}{Gray Hair}}  &  {\rotatebox[origin=c]{90}{Heavy Makeup}}  &  {\rotatebox[origin=c]{90}{High Cheekbones}}  & \\ 
\hline 
PANDA (CVPR14) &88.0 & 78.0 & 81.0 & 79.0 & 96.0 & 92.0 & 67.0 & 75.0 & 85.0 & 93.0 & 86.0 & 77.0 & 86.0 & 86.0 & 88.0 & 98.0 & 93.0 & 94.0 & 90.0 & 86.0 & \\ 
LNets+ANet (ICCV15) &91.0 & 79.0 & 81.0 & 79.0 & 98.0 & 95.0 & 68.0 & 78.0 & 88.0 & 95.0 & 84.0 & 80.0 & 90.0 & 91.0 & 92.0 & 99.0 & 95.0 & 97.0 & 90.0 & 87.0 & \\ 
CTS-CNN (ICB16) &89.0 & 83.0 & 82.0 & 79.0 & 96.0 & 94.0 & 70.0 & 79.0 & 87.0 & 93.0 & 87.0 & 79.0 & 87.0 & 88.0 & 89.0 & 99.0 & 94.0 & 95.0 & 91.0 & 87.0 & \\ 
MT-RBM PCA (CVPRW16) &90.0 & 77.0 & 76.0 & 81.0 & 98.0 & 88.0 & 69.0 & 81.0 & 76.0 & 91.0 & 95.0 & 83.0 & 88.0 & 95.0 & 96.0 & 96.0 & 96.0 & 97.0 & 85.0 & 83.0 & \\ 
Walk and Learn (CVPR16) &84.0 & {\textbf{87.0}} & 84.0 & 87.0 & 92.0 & 96.0 & 78.0 & 91.0 & 84.0 & 92.0 & 91.0 & 81.0 & {\textbf{93.0}} & 89.0 & 93.0 & 97.0 & 92.0 & 95.0 & {\textbf{96.0}} & {\textbf{95.0}} & \\ 
\hline 
MCNN-AUX (AAAI17) &94.5 & 83.4 & 83.1 & 84.9 & 98.9 & {\textbf{96.0}} & 71.5 & 84.5 & {\textbf{89.8}} & {\textbf{96.0}} & 96.2 & 89.2 & 92.8 & 95.7 & 96.3 & {\textbf{99.6}} & 97.2 & {\textbf{98.2}} & 91.5 & 87.6 & \\ 
DMTL (TPAMI17) &{\textbf{95.0}} & 86.0 & {\textbf{85.0}} & {\textbf{99.0}} & 99.0 & 96.0 & {\textbf{88.0}} & {\textbf{92.0}} & 85.0 & 91.0 & 96.0 & {\textbf{96.0}} & 85.0 & {\textbf{97.0}} & {\textbf{99.0}} & 99.0 & {\textbf{98.0}} & 96.0 & 92.0 & 88.0 & \\ 
R-Codean (PRL18) &92.9 & 81.6 & 79.7 & 83.2 & {\textbf{99.5}} & 94.5 & 79.9 & 83.7 & 84.8 & 95.0 & {\textbf{96.6}} & 83.0 & 91.4 & 95.5 & 96.5 & 98.2 & 96.8 & 97.9 & 89.7 & 86.7 & \\ 
Face-SSD &92.9 & 82.0 & 81.3 & 82.5 & 98.6 & 95.2 & 77.8 & 82.3 & 87.9 & 93.6 & 95.0 & 83.5 & 89.6 & 95.1 & 96.0 & 99.2 & 96.3 & 97.6 & 90.7 & 86.8 & \\ 
\hline 
\hline 
 &  {\rotatebox[origin=c]{90}{Male}} &  {\rotatebox[origin=c]{90}{Mouth Slightly Open}} &  {\rotatebox[origin=c]{90}{Mustache}} &  {\rotatebox[origin=c]{90}{Narrow Eyes}} &  {\rotatebox[origin=c]{90}{No Beard}} &  {\rotatebox[origin=c]{90}{Oval Face}} &  {\rotatebox[origin=c]{90}{Pale Skin}} &  {\rotatebox[origin=c]{90}{Pointy Nose}} &  {\rotatebox[origin=c]{90}{Receding Hairline}} &  {\rotatebox[origin=c]{90}{Rosy Cheeks}} &  {\rotatebox[origin=c]{90}{Sideburns}} &  {\rotatebox[origin=c]{90}{Smiling}} &  {\rotatebox[origin=c]{90}{Straight Hair}} &  {\rotatebox[origin=c]{90}{Wavy Hair}} &  {\rotatebox[origin=c]{90}{Wearing Earrings}} &  {\rotatebox[origin=c]{90}{Wearing Hat}} &  {\rotatebox[origin=c]{90}{Wearing Lipstick}} &  {\rotatebox[origin=c]{90}{Wearing Necklace}} &  {\rotatebox[origin=c]{90}{Wearing Necktie}} &  {\rotatebox[origin=c]{90}{Young}} & Mean \\ 
\hline 
PANDA (CVPR14) &97.0 & 93.0 & 93.0 & 84.0 & 93.0 & 65.0 & 91.0 & 71.0 & 85.0 & 87.0 & 93.0 & 92.0 & 69.0 & 77.0 & 78.0 & 96.0 & 93.0 & 67.0 & 91.0 & 84.0 & 85.42 \\ 
LNets+ANet (ICCV15) &98.0 & 92.0 & 95.0 & 81.0 & 95.0 & 66.0 & 91.0 & 72.0 & 89.0 & 90.0 & 96.0 & 92.0 & 73.0 & 80.0 & 82.0 & 99.0 & 93.0 & 71.0 & 93.0 & 87.0 & 87.30 \\ 
CTS-CNN (ICB16) &{\textbf{99.0}} & 92.0 & 93.0 & 78.0 & 94.0 & 67.0 & 85.0 & 73.0 & 87.0 & 88.0 & 95.0 & 92.0 & 73.0 & 79.0 & 82.0 & 96.0 & 93.0 & 73.0 & 91.0 & 86.0 & 86.60 \\ 
MT-RBM PCA (CVPRW16) &90.0 & 82.0 & {\textbf{97.0}} & 86.0 & 90.0 & 73.0 & 96.0 & 73.0 & 92.0 & 94.0 & 96.0 & 88.0 & 80.0 & 72.0 & 81.0 & 97.0 & 89.0 & 87.0 & 94.0 & 81.0 & 86.97 \\ 
Walk and Learn (CVPR16) &96.0 & {\textbf{97.0}} & 90.0 & 79.0 & 90.0 & {\textbf{79.0}} & 85.0 & 77.0 & 84.0 & {\textbf{96.0}} & 92.0 & {\textbf{98.0}} & 75.0 & 85.0 & {\textbf{91.0}} & 96.0 & 92.0 & 77.0 & 84.0 & 86.0 & 88.65 \\ 
\hline 
MCNN-AUX (AAAI17) &98.2 & 93.7 & 96.9 & 87.2 & 96.0 & 75.8 & {\textbf{97.0}} & 77.5 & 93.8 & 95.2 & 97.8 & 92.7 & 83.6 & 83.9 & 90.4 & {\textbf{99.0}} & {\textbf{94.1}} & 86.6 & 96.5 & 88.5 & 91.29 \\ 
DMTL (TPAMI17) &98.0 & 94.0 & 97.0 & 90.0 & {\textbf{97.0}} & 78.0 & 97.0 & {\textbf{78.0}} & {\textbf{94.0}} & 96.0 & {\textbf{98.0}} & 94.0 & {\textbf{85.0}} & {\textbf{87.0}} & 91.0 & 99.0 & 93.0 & 89.0 & {\textbf{97.0}} & {\textbf{90.0}} & {\textbf{92.60}} \\ 
R-Codean (PRL18) &95.9 & 89.8 & 96.3 & {\textbf{90.6}} & 94.6 & 76.5 & 96.9 & 77.0 & 93.6 & 95.3 & 97.6 & 92.8 & 81.2 & 75.4 & 82.7 & 97.9 & 92.0 & {\textbf{89.8}} & 95.9 & 86.6 & 90.14 \\ 
Face-SSD &97.3 & 91.9 & 96.0 & 89.0 & 94.9 & 74.8 & 95.7 & 74.9 & 93.1 & 94.3 & 96.6 & 91.8 & 83.4 & 85.1 & 86.9 & 98.5 & 92.6 & 87.8 & 95.6 & 87.6 & 90.29 \\ 
\hline 
\end{tabular}
\end{table*}

\subsubsection{Facial Attribute Prediction}
\label{subsec: attr prediction performance}

In this section, we evaluated the performance of attribute prediction using Face-SSD for the prediction of $40$ attributes such as gender, age, etc. Our framework treats this problem as multiple binary classification problems using $40$ heatmaps at the output layers. The only difference with the smile recognition case is the number of filter kernels used at the final layer -- everything else remains the same, including the learning hyperparameters. The effects of modifying various settings during training are presented in Table \ref{table: att_color_HaS_Effects}.

Our experiment focuses specifically on the effects of using the Gamma Correction (GC) and Hide-and-Seek (H-a-S) strategies used in the data augmentation process. Depending on the attribute label, there are two possible data augmentation strategies that might affect the accuracy of facial attribute prediction. Gamma correction (colour value adjustment) affects the accuracy of predicting colour-related attributes, such as hair colour (e.g., Black, Blond, Brown and Gray Hair), skin colour (e.g., Pale Skin and Rosy Cheeks) and presence of cosmetics (e.g., Heavy Makeup and Wearing Lipstick). Hide-and-Seek, which forces the Face-SSD to seek more of the overall face area, seems to affect the accuracy of predicting the overall face area-related attributes including ``Attractive, Blurry, Chubby, Heavy Makeup, Oval Face, Pale Skin and Young''.

\begin{figure} [t!]
\small
\begin{center}
      \includegraphics[width=0.95\linewidth]{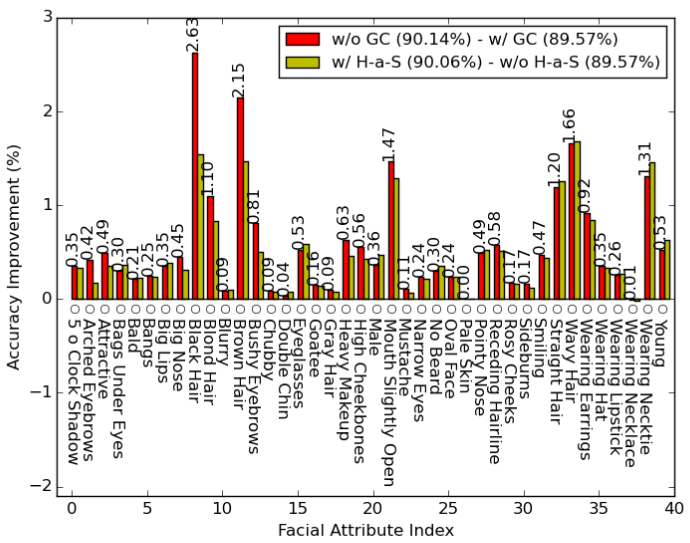}
\end{center}
\caption{{Removing Gamma Correction (GC) during training Face-SSD (Case C in Table \ref{table: att_color_HaS_Effects}) improves the accuracy of predicting color-related attributes comparing to using GC (Case A in Table \ref{table: att_color_HaS_Effects}). Using Hide-and-Seek (H-a-S) (Case B in Table \ref{table: att_color_HaS_Effects}) does not improve overall face area-related attributes as expected.}}
\label{fig:effect of using GC and H-a-S}
\end{figure}

As shown in Table \ref{table: att_color_HaS_Effects}, we tested Face-SSD with all possible combinations using Gamma Correction and Hide-and-Seek during training, and all other settings remained the same as face detection part in Face-SSD (See Table \ref{table: FD_HNM_HaS_Effects}). As we expected, using Gamma Correction (Case A and B in Table \ref{table: att_color_HaS_Effects}), which modifies the original colour of the training image, degrades the attribute recognition performance compared to training without Gamma Correction (Case C and D in Table \ref{table: att_color_HaS_Effects}). Although training without Gamma Correction primarily improves the accuracy of the colour-related attributes (e.g., Black Hair, Blond Hair, Brown Hair and Heavy Makeup), it also helps improve overall accuracy in other attributes, as shown in Fig. \ref{fig:effect of using GC and H-a-S}. By removing only Gamma Correction, Face-SSD achieves state-of-the-art accuracy ($90.29\%$) that is competitive results ($> 90\%$) similarly to MCNN-AUX {\citep{conf/AAAI/Hand17}}, DMTL {\citep{jour/PAMI/Han17}} and R-Codean {\citep{jour/PRL/Sethi18}}. (See Fig. \ref{fig:ACC_multi_att})

Interestingly, the use of Hide-and-Seek improves accuracy, but does not primarily improve the accuracy of attributes that are related to large facial areas, such as ``Attractive, Blurry, Chubby, Heavy Makeup, Oval Face, Pale Skin and Young'' as it was originally expected. On the contrary, it helps to identify more details in certain face areas (e.g., Bushy Eyebrows, Mouth Slightly Open, Straight Hair, Wavy Hair, Wearing Earrings, Wearing Necktie), as shown in Fig. \ref{fig:effect of using GC and H-a-S}. When training without using Gamma Correction, Face-SSD does not benefit from the use of Hide-and-Seek, as shown in Table \ref{table: att_color_HaS_Effects} (Case D). The reason for this is that training without using Gamma Correction has had more impact on improving the accuracy of the same attributes, as shown in Fig. \ref{fig:effect of using GC and H-a-S}. The results of the Face-SSD shown in Table \ref{table: comparison_att_CelebA} are obtained by training Face-SSD using of Hide-and-Seek (Case D in Table \ref{table: att_color_HaS_Effects}), but not using Gamma Correction. Although we use the generalised Face-SSD architecture as opposed to using a specially designed architecture for facial attribute prediction, we achieved state-of-the-art accuracy (in the top three among the performances of related works).

\begin{table}[!t]
\small
\caption{The effect of using Gamma Correction (GC) and Hide-and-Seek (H-a-S) in the data augmentation process when training Face-SSD for attribute prediction using CelebA dataset.}
\label{table: att_color_HaS_Effects}
\centering
\begin{tabular}{|c|c|c||c|}
\hline
 						& Using GC
& Using H-a-S
& Accuracy ($\%$) \\
\hline\hline
Face-SSD A
& $\checkmark$			& $\cdot$
& 89.57	\\
\hline
Face-SSD B
& $\checkmark$			& $\checkmark$
& 90.06		\\
\hline
Face-SSD C
& $\cdot$				& $\cdot$
& 90.15 	\\
\hline
Face-SSD D
& $\cdot$				& $\checkmark$
& 90.29	\\
\hline

\end{tabular}
\end{table}

\subsubsection{Valence and Arousal Estimation}
\label{subsec: V-A estimation performance}

In this section, we investigated the performance of valence-arousal estimation using Face-SSD. Unlike the previous sections that address binary classification (smile recognition) and multi-class recognition (facial attribute prediction) problems, Face-SSD for valence-arousal solves a regression problem. To this end we used a state-of-the-art dataset called AffectNet {\citep{jour/TAC/Mollahosseini17}}. AffectNet consists of face images captured in the wild and its corresponding annotations of valence-arousal and emotion. To confirm the regression ability of Face-SSD, we only investigated the valence-arousal estimation performance.

Note that, as AffectNet consists only of cropped face images, we trained Face-SSD using a data augmentation strategy that allows only minor variations in terms of face size. Therefore, during testing, Face-SSD typically handles large faces for valence-arousal estimation. Despite this limitation during training, however, Face-SSD is able to handle not only large faces but also faces of medium size during testing, as shown in Fig. \ref{fig:teaser}(c).

The performance of the valence-arousal estimation is shown in Table \ref{table:ACC_val_aro}. For valence estimation, AffectNet yields slightly better results than Face-SSD. On the other hand, in terms of arousal, Face-SSD provides better results. Overall, Face-SSD provides close to the state-of-the-art performance without any modification to the original architecture of the Face-SSD network. See {\citep{jour/TAC/Mollahosseini17}} for a detailed description of the units in the Table \ref{table:ACC_val_aro}.

\begin{table}[!t]
\small
\caption{Experimental results of valence and arousal estimation using AffectNet {\citep{jour/TAC/Mollahosseini17}} dataset. Experimental results are reported using Root Mean Square Error (RMSE), Pearsons Correlation Coefficient (CoRR), Sign Agreement Metric (SAGR) and Concordance Correlation Coefficient (CCC) (see {\citep{jour/TAC/Mollahosseini17}} for the detailed description of the metrics).}
\label{table:ACC_val_aro}
\centering
\begin{tabular}{|c||c|c|c|c|}
\hline
			&
\multicolumn{2}{ |c| }{Valence} & 
\multicolumn{2}{ |c| }{Arousal}\\
\cline{2-5}
     		&
AffectNet	& Face-SSD	&
AffectNet	& Face-SSD \\
\hline\hline
RMSE 		&
{\textbf{0.37}}	& 0.4406	&
0.41			& {\textbf{0.3937}}	\\
\hline
CORR 		&
{\textbf{0.66}}	& 0.5750 	& {\textbf{0.54}} & 0.4953	\\
\hline
SAGR 		&
{\textbf{0.74}}	& 0.7284	&
0.65			& {\textbf{0.7129}}	\\
\hline
CCC 		&
{\textbf{0.60}}	& 0.5701	&
  0.34			& {\textbf{0.4665}} \\
\hline
\end{tabular}
\end{table}

\subsection{Computational Speed and Complexity}
\label{subsec: computation and complexity}
For all of the Face-SSD applications presented in this paper, we obtained an average processing time of ${\textbf{47.39}}$ $ms$ (${\textbf{21.10}}$ $FPS$) during testing, with an experimental environment consisting of an Intel Core i7-6700HQ CPU processor and an NVIDIA GeForce GTX 960M GPU with 23.5GB of DRAM. We used Theano for Face-SSD implementation. As shown in Table \ref{table:proc_time_param_cnt}, most Face-SSD applications achieve near real-time processing speed. Smile recognition (binary classification), facial attribute prediction (40-class recognition) and valence-arousal estimation (multiple task regression) take $47.28$ $ms$ ($21.15$ $FPS$), $47.55$ $ms$ ($21.03$ $FPS$) and $47.37$ $ms$ ($21.11$ $FPS$), respectively. Using the proposed generic Face-SSD for face analysis, the number of model parameters indicating complexity does not increase linearly even when the number of facial analysis tasks and classes increases. Although facial attribute prediction performs $40$ times more tasks than smile recognition, the processing time by the attribute prediction task increases only by $0.27$ $ms$ and requires a small number of additional parameters ($0.09$ $M$).

As shown in Table \ref{table: comparison_CelebA}, the proposed Face-SSD is significantly faster than traditional methods that use the steps of region proposal and task prediction to analyse faces. For example, the work of Liu et al. {\citep{conf/ICCV/Liu15}} requires $35$ $ms$ to generate the face confidence heatmap and $14$ $ms$ to classify the attributes. In addition, this method requires another $90$ $ms$ to find the candidate bounding box (EdgeBox {\citep{conf/ECCV/Zitnick14}}) for localising the final bounding box that ends up with a total processing time of $139$ $ms$ ($7.19$ $FPS$). The work of Ranjan et al. {\citep{conf/FG/Ranjan16}} takes an average of $3,500$ $ms$ ($0.29$ $FPS$) to process an image. Ranjan et al. {\citep{conf/FG/Ranjan16}} explains that the main bottleneck for speed is the process of proposing regions (Selective Search {\citep{conf/ICCV/Sande11}}) and the repetitive CNN process for every individual proposal.

\begin{table}[!t]
\small
\caption{The total number of parameters and processing time for various face analysis applications using Face-SSD.}
\label{table:proc_time_param_cnt}
\centering
\begin{tabular}{|c|c|c|}
\hline
Face Analysis Task 			& Parameter Number 	& ms (FPS) \\
\hline\hline
Face Detection Part (only)	& $2.31$ $M$ 
& 25.57 (39.11) \\
\hline
\hline
Smile Recognition			& $4.44$ $M$ 
& 47.28 (21.15) \\
\hline
Facial Attribute Prediction	& $4.53$ $M$ 
& 47.55 (21.03) \\
\hline
Valence-Arousal Estimation	& $4.46$ $M$ 
& 47.37 (21.11) \\
\hline
\hline
Average of All Applications	& $4.48$ $M$ 
& 47.39 (21.10) \\
\hline
\end{tabular}
\end{table}

To ensure a fair comparison of the processing times, we should measure the time in the same experimental environment. However, Liu et al. {\citep{conf/ICCV/Liu15}} does not provide detailed information about the experimental environment, except that they use GPUs. Ranjan et al. {\citep{conf/FG/Ranjan16}} implemented their all-in-one network using $8$ CPU cores and GTX TITAN-X GPUs. The processing speed of the proposed Face-SSD is $74$ times faster than the all-in-one network, even in a less powerful experimental environment.

Although Face-SSD is faster than other face analysis methods, the processing speed is lower than the base object detection (SSD) model {\citep{conf/ECCV/Liu16}} as the complexity of Face-SSD is nearly twice that of SSD, as shown in Table \ref{table:proc_time_param_cnt}. Placing more layers to perform face analysis tasks increased the number of parameters in Face-SSD. However, the structure of the all-in-one network {\citep{conf/FG/Ranjan16}} shows that sharing more convolutional features does not degrade the performance of various tasks. Capitalising on this idea, we expect to further reduce the complexity of Face-SSD by sharing more layers and assigning a relatively small number of layers to other face analysis tasks.

\section{Conclusions}
\label{sec: conclusion}
In this paper, we tackled the problem of multiple face analysis tasks, namely smile recognition, facial attribute prediction and valence-arousal estimation in the wild, without the traditional pre-normalisation steps of face detection and registration. To this end, we proposed Face-SSD which performs face detection and face analysis simultaneously in a single framework. For fast and scale-invariant detection, Face-SSD builds upon the state-of-the-art object detection network SSD. In addition, we used pre-trained parameters of two different networks, trained for object classification and for face detection, to learn the face and task-relevant patterns. Consequently, we built a single framework that enables real-time scale-invariant face analysis in the wild. By exploring various data augmentation strategies for face analysis while maintaining the same Face-SSD architecture, we achieved state-of-the-art performance for various face analysis tasks without increasing model complexity. Our experimental results show that Face-SSD achieves state-of-the-art performance (accuracy of $95.76\%$ for smile recognition, and $90.29\%$ for attribute prediction, RMSE of $0.44$ and $0.39$ for valence and arousal estimation) while maintaining real-time speed ($21.15$ $FPS$ for smile recognition, $21.03$ $FPS$ for attribute prediction, $21.11$ $FPS$ for valence-arousal estimation). For our future work, we plan to investigate a way of using facial attributes to improve the face detection performance. The challenge for doing this involves using heterogeneous annotations contained in separate datasets.

\section*{Acknowledgments}
This work has been supported by the Technology Strategy Board, UK / Innovate UK project Sensing Feeling (project no. 102547). This work was undertaken while Youngkyoon Jang was a research associate affiliated with Queen Mary University of London and University of Cambridge.

\bibliographystyle{model2-names}
\bibliography{./CVIU_Face_SSD_review.bbl}

\end{document}